\documentclass[letterpaper]{article}
\usepackage{chicagor}
\usepackage{aaai1}
\usepackage{times}
\usepackage{helvet}
\usepackage{courier}
\newtheorem{THEOREM}{Theorem}[section]
\newenvironment{theorem}{\begin{THEOREM} \hspace{-.85em} {\bf :} }%
                        {\end{THEOREM}}
\newtheorem{LEMMA}[THEOREM]{Lemma}
\newenvironment{lemma}{\begin{LEMMA} \hspace{-.85em} {\bf :} }%
                      {\end{LEMMA}}
\newtheorem{COROLLARY}[THEOREM]{Corollary}
\newenvironment{corollary}{\begin{COROLLARY} \hspace{-.85em} {\bf :} }%
                          {\end{COROLLARY}}
\newtheorem{PROPOSITION}[THEOREM]{Proposition}
\newenvironment{proposition}{\begin{PROPOSITION} \hspace{-.85em} {\bf :} }%
                            {\end{PROPOSITION}}
\newtheorem{DEFINITION}[THEOREM]{Definition}
\newenvironment{definition}{\begin{DEFINITION} \hspace{-.85em} {\bf :} \rm}%
                            {\end{DEFINITION}}
\newtheorem{CLAIM}[THEOREM]{Claim}
\newenvironment{claim}{\begin{CLAIM} \hspace{-.85em} {\bf :} \rm}%
                            {\end{CLAIM}}
\newtheorem{EXAMPLE}[THEOREM]{Example}
\newenvironment{example}{\begin{EXAMPLE} \hspace{-.85em} {\bf :} \rm}%
                            {\end{EXAMPLE}}
\newtheorem{REMARK}[THEOREM]{Remark}
\newenvironment{remark}{\begin{REMARK} \hspace{-.85em} {\bf :} \rm}%
                            {\end{REMARK}}
\newcommand{\thm}{\begin{theorem}}
\newcommand{\lem}{\begin{lemma}}
\newcommand{\pro}{\begin{proposition}}
\newcommand{\dfn}{\begin{definition}}
\newcommand{\rem}{\begin{remark}}
\newcommand{\xam}{\begin{example}}
\newcommand{\cor}{\begin{corollary}}
\newcommand{\prf}{\noindent{\bf Proof:} }
\newcommand{\ethm}{\end{theorem}}
\newcommand{\elem}{\end{lemma}}
\newcommand{\epro}{\end{proposition}}
\newcommand{\edfn}{\bbox\end{definition}}
\newcommand{\erem}{\bbox\end{remark}}
\newcommand{\exam}{\bbox\end{example}}
\newcommand{\ecor}{\end{corollary}}
\newcommand{\eprf}{\bbox\vspace{0.1in}}
\newcommand{\beqn}{\begin{equation}}
\newcommand{\eeqn}{\end{equation}}
\newcommand{\bbox}{\vrule height7pt width4pt depth1pt}

\newcommand{\clm}{\begin{claim}}
\newcommand{\eclm}{\end{claim}}
\newcommand{\sat}{\models}

\newcommand{\bor}{\bigvee}
\newcommand{\union}{\cup}
\newcommand{\inter}{\cap}
\newcommand{\FF}{{\bf F}}

\renewcommand{\phi}{\varphi}
\newcommand{\F}{{\cal F}}

\newcommand{\K}{{\cal K}}

\newcommand{\R}{{\cal R}}

\newcommand{\U}{{\cal U}}
\newcommand{\V}{{\cal V}}

\renewcommand{\>}{\rangle}

\newcommand{\ol}{\setlength{\itemsep}{0pt}\begin{enumerate}}
\newcommand{\eol}{\end{enumerate}\setlength{\itemsep}{-\parsep}}
\newcommand{\ul}{\setlength{\itemsep}{0pt}\begin{itemize}}
\newcommand{\dl}{\setlength{\itemsep}{0pt}\begin{description}}
\newcommand{\edl}{\end{description}\setlength{\itemsep}{-\parsep}}
\newcommand{\eul}{\end{itemize}\setlength{\itemsep}{-\parsep}}

\newcommand{\BS}{B^{\scriptscriptstyle \cS}}

\newcommand{\commentout}[1]{}

\newcommand{\bi}{\begin{itemize}}
\newcommand{\ei}{\end{itemize}}
\newcommand{\be}{\begin{enumerate}}
\newcommand{\ee}{\end{enumerate}}

\newenvironment{oldthm}[1]{\par\noindent{\bf Theorem #1:} \em \noindent}{\par}
\newenvironment{oldlem}[1]{\par\noindent{\bf Lemma #1:} \em \noindent}{\par}
\newenvironment{oldcor}[1]{\par\noindent{\bf Corollary #1:} \em \noindent}{\par}
\newenvironment{oldpro}[1]{\par\noindent{\bf Proposition #1:} \em \noindent}{\par}
\newcommand{\othm}[1]{\begin{oldthm}{\ref{#1}}}
\newcommand{\eothm}{\end{oldthm} \medskip}
\newcommand{\olem}[1]{\begin{oldlem}{\ref{#1}}}
\newcommand{\eolem}{\end{oldlem} \medskip}
\newcommand{\ocor}[1]{\begin{oldcor}{\ref{#1}}}
\newcommand{\eocor}{\end{oldcor} \medskip}
\newcommand{\opro}[1]{\begin{oldpro}{\ref{#1}}}
\newcommand{\eopro}{\end{oldpro} \medskip}

\newcommand{\bxor}[1]{\dot{\bor}}

\renewcommand{\S}{{\cal S}}

\renewcommand{\S}{{\cal S}}
\newcommand{\Suff}{\mathbf{S}}
\newcommand{\ML}{\mathit{M}}

\renewcommand{\FF}{\mathit{FF}}

\newcommand{\VS}{\mathit{VS}}

\newcommand{\WB}{\mathit{WB}}
\newcommand{\BMC}{\mathit{BMC}}
\newcommand{\TT}{\mathit{TT}}

\newcommand{\ST}{\mathit{ST}}

\newcommand{\MT}{\mathit{T}}
\newcommand{\CP}{\mathit{CP}}
\newcommand{\AC}{\mathit{DAP}}
\newcommand{\PD}{\mathit{PD}}

\newcommand{\BH}{\mathit{BH}}
\renewcommand{\BS}{\mathit{BS}}
\newcommand{\BT}{\mathit{BT}}
\newcommand{\SH}{\mathit{SH}}

\newcommand{\dr}{\mbox{{\em dr}}}
\newcommand{\db}{\mbox{{\em db}}}
\newcommand{\WIN}{\mbox{{\em WIN}}}
\renewcommand{\gets}{=}
\newcommand{\fullv}[1]{#1}
\newcommand{\shortv}{\commentout}
\newcommand{\kr}[1]{#1}
\newcommand{\nokr}{\commentout}
\newcommand{\journal}{\commentout}
\shortv{\nocopyright}
\renewcommand{\chicagoraddresspub}{\commentout}

\begin{document}

\journal{\begin{titlepage}}
\shortv{\setlength\titlebox{2.0in}}
\title{Defaults and Normality in Causal Structures}
\author{Joseph Y. Halpern%
\fullv{\thanks{Supported in part by NSF under
under grants ITR-0325453 and IIS-0534064, and by AFOSR under grant 
FA9550-05-1-0055.}}\\
Cornell University\\
Dept. of Computer Science\\
Ithaca, NY 14853\\
halpern@cs.cornell.edu\\
http://www.cs.cornell.edu/home/halpern}

%\author{\mbox{ \ \ \ }}

%\date{ }

\maketitle

\fullv{\setcounter{page}{0}
\thispagestyle{empty}
}
%\end{fullv}
%\begin{document}
%\thispagestyle{empty}

\begin{abstract}
A serious defect with the Halpern-Pearl (HP) definition of
causality is repaired by combining a theory of causality with a theory
of defaults.  In addition, it is shown that (despite a
claim to the contrary) a cause according to the HP condition need not be
a single conjunct.  A definition of causality motivated
by Wright's NESS test is shown to always hold for a
single conjunct.  Moreover, conditions that hold for all the examples
considered by HP are given that guarantee
that causality according to (this version) of the NESS test is
equivalent to the HP definition.
\end{abstract}

\journal{\end{titlepage}}

\section{Introduction}
Getting an adequate definition of causality is difficult.  There have
been numerous attempts, in fields ranging from philosophy 
to law to computer science (see, e.g., \cite{Collins03,HH85,pearl:2k}).
A recent definition by Halpern and Pearl (HP from now on), first
introduced in \cite{HPearl01a}, using structural equations, has
attracted some attention recently.  The intuition behind this
definition, which goes back to Hume \citeyear{hume:1748}, is
that $A$ is a cause of $B$ if, had $A$ not happened, $B$ would not have
happened.  
%joe1
For example, despite the fact that it was raining and I was
drunk, the faulty brakes are the cause of my accident because, had the
brakes not been faulty, I would not have had the accident.
As is well known, 
this definition does not quite work.  
\commentout{
To take an example due to Hall
\citeyear{Hall98}, suppose that Billy 
and Suzy both throw rocks at a bottle, but Suzy's rock hits first,
shattering the bottle. We would like to call Suzy's throw a cause of the
bottle shattering, but if Suzy hadn't thrown, Billy's rock would have
hit, and the bottle would have shattered anyway.  HP deal with this by,
roughly speaking, considering the contingency where Billy does not
throw.  Under that contingency, the bottle shatters if Suzy throws, and
does not shatter if she does not throw.  To stop Billy's throw from also
being a cause, HP put some constraints on the contingencies that could be
considered.}
To take an example due to Wright \citeyear{Wright85}, suppose that
Victoria, the victim, 
drinks a cup of tea poisoned by Paula, but before the poison takes
effect, Sharon shoots Victoria, and she dies.
We would like to call Sharon's shot the cause of 
%joe1
the
Victoria's death, but if Sharon hadn't shot, Victoria would have died in any
case. HP deal with this by,
roughly speaking, considering the contingency where Sharon does not
shoot.  Under that contingency, Victoria dies if Paula administers
the poison, and otherwise does not.  To prevent the poisoning from also
being a cause of Paula's death, HP put some constraints on the
contingencies that could be considered.

Unfortunately, two significant problems have been found with the
original HP definition, each leading to situations where the definition does
not match most people's intuitions regarding causality.
The first, observed by Hopkins and
Pearl \citeyear{HopkinsP02} (see Example~\ref{xam3}), showed that 
the constraints on the contingencies were too liberal.  This problem
was dealt with in the journal version of the HP paper \cite{HP01b} by
putting a further constraint on contingencies.
The second problem is arguably deeper.  As examples of Hall \citeyear{Hall07}
and Hiddleston \citeyear{Hiddleston05} show, the HP definition gives
inappropriate answers in cases that have structural equations isomorphic
to ones where the HP definition gives the appropriate answer (see
Example~\ref{xam:bogus}).  Thus, there must be more to causality than
just the structural equations.  The final HP definition recognizes this
problem by viewing some contingencies as ``unreasonable'' or
``farfetched''.  However, in some of the examples, it
is not clear why the relevant contingencies are more farfetched than
others.  I show that the problem is even deeper than that: there is
no way of viewing contingencies as ``farfetched'' independent of
actual contingency that can solve the problem.

This paper has two broad themes, motivated by the two problems in the HP
definition.  First, 
I propose a general approach for dealing with the second problem,
%determining what counts as reasonable, which is 
motivated by the following well-known observation
in the psychology literature \cite[p.~143]{KM86}:
``an event is more likely to be undone by altering exceptional than
routine aspects of the causal chain that led to it.''  In the language of
this paper, a contingency that differs from the actual situation by
changing something that is atypical in the actual situation is more
reasonable than one that differs by changing something that is typical
in the actual situation.  
\journal{As Kahnemann and Miller point out, this
intuition is also present in the legal literature.  Hart and Honor{\'e}
\citeyear{HH85} observe that the statement ``It was the presence of
oxygen that caused the fire'' makes sense only if there were reasons to
view the presence of oxygen as abnormal.  
}
%\end{fullv}
%Putting this in the HP
%framework, that means we do not want to consider 
%a contingency where there is no oxygen, even if the presence of
%oxygen is an endogenous variable, unless there is reason to take the
%absence of oxygen as normal (for example, at the top of Mount Everest).
To capture this intuition formally, I use a well-understood
approach to dealing with defaults and normality \cite{KLM}.  
Combining a default theory with causality, using the intuitions 
of Kahnemann and Miller,  leads to a straightforward solution to the
second problem.  The idea is that, when showing that if $A$
hadn't happened then $B$ would not have happened, we consider only
contingencies that are more normal than the actual world.
For example, if someone typically leaves work at 5:30 PM and arrives home at
6, but, due to unusually bad traffic, arrives home at 6:10, the bad
traffic is typically viewed as the cause of his being late, not 
the fact that he left at 5:30 (rather than 5:20).

%While combining default reasoning and causal reasoning is the main
%contribution of this paper, the examination of the HP solution to the
%problem pointed out by Hopkins and Pearl also leads to further
%insights.

The second theme of this paper is a comparison of the HP definition to 
perhaps the best worked-out approach to causality in the
legal literature: the NESS (Necessary Element of a Sufficient Set) test,
originally described by Hart and Honor\'{e} \citeyear{HH85}, and worked
out in \nokr{much} greater detail by Wright
\citeyear{Wright85,wright:88,Wright01}.   This 
%joe1
is motivated
in part by the
first problem.
%Going back to the first problem, while the solution to this problem
%proposed in \cite{HP01b} does deal with it,
%%with the problem, 
%it introduces a new (smaller) problem.  
As shown by
Eiter and Lukasiewicz \citeyear{EL01} and Hopkins
\citeyear{Hopkins01},
the original HP definition had the property that causes were always
single conjuncts; that is, it is never the case that $A \land A'$ is a
cause of $B$ if $A \ne A'$. 
%(for the class of formulas considered by HP).  
This property, which plays a critical role in the complexity
results of Eiter and Lukasiewicz \citeyear{EL01}, was also claimed to
hold for the revised definition  
\cite{HP01b} (which was revised precisely to deal with the first problem)
but, as I show here, it does not.  Nevertheless, for all
the examples considered in the literature, the cause is always a single
conjunct.  
%One (somewhat roundabout) way of understanding why is given
%by considering  the NESS test.
Considering the NESS test helps explain why.
\journal{[[JOE:]] put in a footnote. this is different from the issue
that there are typically many causes, and we tend to focus on one.  I
return to this issue in the Discussion.}

While the NESS test is simple and intuitive, and deals well with many
examples, as I show here, it suffers from some serious problems.  In
In particular, it lacks a clear definition of what it means for a set of
events to be \emph{sufficient} for another event to occur.  I provide
such a definition here, using ideas from the HP definition of
causality.  Combining these ideas with the intuition behind the NESS
test leads to a definition of causality that (a) often agrees with the
HP definition (indeed, does so on all the examples in the HP paper) and
(b) has the property that a cause is always a single conjunct.  I
provide a sufficient condition (that holds in all the examples in the HP
paper) for when the NESS test definition implies
the HP definition, thus also providing an explanation as to why the
cause is a single conjunct according to the HP definition in so many cases.

I conclude this introduction with a brief discussion on related work.
There has been a great deal of work on causality in philosophy,
statistics, AI, and the law.  It is beyond the scope of this paper to
review it; the HP paper has some comparison of the HP approach to
other, particularly those in the philosophy literature.  It is perhaps
worth mentioning here that the focus of this work is quite different
from the AI work on 
formal action theory (see, for example,
\cite{lin:95,sandewall:94,reiter:01}), 
which is concerned with
applying causal relationships so as to guide actions, 
as opposed to the focus here on extracting the actual causality relation
from a specific scenario.

\nokr{
The rest of this paper is organized as follows.  
\fullv{
In Section~\ref{sec:causalmodel}, I provide a brief introduction to
structural equations and causal models, so as to make this paper
self-contained. 
In Section~\ref{sec:actcaus}, I review the HP definition, and show that,
in general, causes are not always single conjuncts.}
\shortv{
In Section~\ref{sec:causalmodel}, I provide a brief introduction to
structural equations, causal models, and the HP definition, so as to
make this paper self-contained. 
I also show that, in general, causes are not always single conjuncts.}
In Section~\ref{sec:final}, I show how the HP definition
can be combined with standard approaches for modeling defaults, and how
that deals with the various problems that have been raised.
%In Section~\ref{sec:actcaus}, I
%review the HP definition of actual causality.  
%Section~\ref{sec:responsibility}, I show how the definition can be
%extended to capture reasponsibility and blame.  
In Section~\ref{sec:NESS}, I compare the
structural-model definition of causality is compared to Wright's
\citeyear{Wright85,wright:88,Wright01} NESS test, and give 
a formal analogue of the NESS test combined with ideas in the HP
definition.  I conclude in
Section~\ref{sec:conc}.
\kr{Proofs are left to the full paper.}
\nokr{Proofs can be found in the appendix.}  
%The material in Sections~\ref{sec:causalmodel}, \ref{sec:actcaus}, and
%\ref{sec:responsibility} is largely taken 
%from \cite{HP01b} and \cite{ChocklerH03}; the reader is encouraged to
%consult these papers for more details, more intuition, and more
%examples.  
}

\shortv{\section{The HP Definition of Causality}\label{sec:causalmodel}}
\fullv{\section{Causal Models}\label{sec:causalmodel}}

In this section, I briefly review the formal model of causality used in
the HP definition.  
%The description of causal models given here is taken
%from \cite{Hal20}; it is a formalization of earlier work of Pearl
%\citeyear{Pearl.Biometrika}, further developed in 
%\cite{GallesPearl97,Hal20,pearl:2k}.  
More details, intuition, and motivation can be found in
\shortv{\cite{HP01b}.}  
\fullv{\cite{HP01b} and the references therein.}

%\paragraph{Causal models:}
The HP approach assumes that the world is described in terms of random
variables and their values.  
For example, if we are trying to determine
whether a forest fire was caused by lightning or an arsonist, we 
can take the world to be described by three random variables:
%\fullv{
%\begin{itemize}
%\item }
$\FF$ for forest fire, where $\FF=1$ if there is a forest fire and
$\FF=0$ otherwise; 
%\fullv{\item}
$L$ for lightning, where $L=1$ if lightning occurred and $L=0$ otherwise;
%\fullv{\item} 
$\ML$ for match (dropped by arsonist), where $\ML=1$ if the arsonist
drops a lit match, and $\ML = 0$ otherwise.
%\fullv{\end{itemize}}
\fullv{
The choice of random variables determines the language used to frame
the situation.  Although there is no ``right'' choice, clearly some
choices are more appropriate than others.  For example, when trying to
determine the cause of Sam's lung cancer, if there is no random variable
corresponding to smoking in a model then, in that model, we cannot hope to
conclude that smoking is a cause of Sam's lung cancer.  

}
%\end{fullv}
Some random variables may have a causal influence on others. This
influence is modeled by a set of {\em structural equations}.
For example, 
%if we want 
to model the fact that
if a match is lit or lightning strikes then a fire starts, we could use
the random variables $\ML$, $\FF$, and $L$ as above, with the equation
$\FF = \max(L,\ML)$. 
%that is, the value of the random variable $\FF$ is the
%maximum of the values of the random variables $\ML$ and $L$.  Thus, for
%example, the equation says that if $\ML=0$ and $L=1$, then $\FF=1$.
\fullv{The equality sign in this equation should be thought of more like an 
assignment statement in programming languages; once we set the values of
$\FF$ and $L$, then the value of $\FF$ is set to their maximum.  However,
despite the equality, if a forest fire starts some other way, that does not
force the value of either $\ML$ or $L$ to be 1.  
}

\journal{Alternatively, if we want to model the fact that a fire
requires both a 
lightning strike \emph{and} a dropped match (perhaps the wood is so wet
that it needs two sources of fire to get going).  The only change in the
model is that the equation for $\FF$ becomes $\FF = \min(L,\ML)$; the
value of $\FF$ is the minimum of the values of $\ML$ and $L$.  The only
way that $\FF = 1$ is if both $L=1$ and $\ML=1$.  
%For future reference,
%call the model of forest fires where either the match or the lightning
%result in fire the \emph{disjunctive} model, and call the model where
%both are required the \emph{conjunctive} model.  

Both of these models are somewhat simplistic.  Lightning does not always
result 
in a fire, nor does dropping a lit match.  One way of dealing with this
would be to make the assignment statements probabilistic.  For example,
we could say that the probability that $\FF=1$ conditional on $L=1$
is .8.  This approach would lead to rather complicated definitions.
It is much simpler to think of all the equations as being deterministic
and, intuitively, use enough variables to capture all the conditions
that determine whether there is a forest fire are captured by
random variables.  One way to do this is simply to add those variables
explicitly.  For example, we could add variables that talk about the
dryness of the wood, the amount of undergrowth, the presence of
sufficient oxygen  (fires do not start so easily on the top of high
mountains), and so on.  If a modeler does not want to add all these
variables explicitly (the details may simply not be relevant to the
analysis), another alternative is to use a single variable,
say $U$, which intuitively incorporates all the relevant factors,
without describing them explicitly.  The value of $U$ would determine
whether the lightning strikes, whether the match is dropped by
the arsonist, and whether both are needed to start a fire, just one, or
neither (perhaps fires start spontaneously, or there is a cause not
modeled by the random variables used).  In this way of modeling things,
$U$ would take on 12 possible values of the form $(i,j,k)$, where $i$
and $j$ are both either 0 or 1 and $k \in \{0,1,2\}$.  Intuitively,
$i$ describes whether the external conditions are such that the
lightning strikes (and encapsulates all the conditions, such as humidity
and temperature, that 
affect whether the lightning strikes); $j$ describes whether 
the arsonist drops the match (and thus encapsulates the psychological
conditions that determine whether the arsonist drops the match); and $k$
describes whether neither, just one, or both a dropped match and
lightning are needed for the lightning to strike.
Thus, the equation the equation could say that $\FF=1$ if, for example,
$U=(1,1,1)$ (so that the lightning strikes, the match is dropped, and
only one of them is needed to have a fire) or if $U = (1,0,1)$, or if 
$U = (1,1,2)$, but not if $U = (1,0,2)$.  
}
%\end{journal}

It is conceptually useful to split the random variables into two
sets: \fullv{the} {\em exogenous\/} variables, whose values are
determined by 
factors outside the model, and \fullv{the}
{\em endogenous\/} variables, whose values are ultimately determined by
the exogenous variables.  For example, in the forest fire example, the
variables $\ML$, $L$, and $\FF$ are endogenous. 
%variable $U$ could be exogenous.  
However, we want to take as given that there is
enough oxygen for the fire and that the wood is sufficiently dry to
burn.  In addition, we do not want to concern ourselves with the factors
that make the arsonist drop the match or the factors that cause lightning.
These factors are all determined by the exogenous variables.  

%the endogenous variables whose values are described by the
%structural equations.  In the forest fire example, the variables
%$\FF$, $L$, and $\ML$ are endogenous.  As we shall see,
%when we want to talk about the probability that $A$ is a cause of $B$,
%we can do so quite easily by putting a probability distribution on the
%values of the exogenous random variables.  This gives us a way of
%talking about the probability of there being a fire if lightning
%strikes, while still using deterministic equations.

%With this background, we can formally define a \emph{causal model} $M$
%to be a pair $(\S,\F)$, where $\S$ is a \emph{signature}, which
%explicitly
Formally, a \emph{causal model} $M$
is a pair $(\S,\F)$, where $\S$ is a \emph{signature}, which explicitly
lists the endogenous and exogenous variables  and characterizes
their possible values, and $\F$ defines a set of \emph{modifiable
structural equations}, relating the values of the variables.  
%In the
%next two paragraphs, i  define $\S$ and $\F$ formally; the definitions
%can be skipped by the less mathematically inclined reader.  
A signature $\S$ is a tuple $(\U,\V,\R)$, where $\U$ is a set of
exogenous variables, $\V$ is a set 
of endogenous variables, and $\R$ associates with every variable $Y \in 
\U \union \V$ a nonempty set $\R(Y)$ of possible values for 
\fullv{$Y$ (that is, the set of values over which $Y$ {\em ranges}).}  
\shortv{$Y$.}
\journal{CHECK:
As suggested above, in the
forest-fire example, we have $\U = \{U\}$, where $U$ is the
exogenous variable, $\R(U)$ consists of the 12 possible values of $U$ 
discussed earlier, $\V = 
\{\FF,L,\ML\}$, and $\R(\FF) = \R(L) = \R(\ML) = \{0,1\}$.  
}
$\F$ associates with each endogenous variable $X \in \V$ a
function denoted $F_X$ such that $F_X: (\times_{U \in \U} \R(U))
\times (\times_{Y \in \V - \{X\}} \R(Y)) \rightarrow \R(X)$.
%\footnote{Recall that $\V - \{X\}$ is the set consisting of all the
%variables in $\V$ that are not in $\{X\}$.}
This mathematical notation just makes precise the fact that 
$F_X$ determines the value of $X$,
given the values of all the other variables in $\U \union \V$.
If there is one exogenous variable $U$ and three endogenous
variables, $X$, $Y$, and $Z$, then $F_X$ defines the values of $X$ in
terms of the values of $Y$, $Z$, and $U$.  For example, we might have 
$F_X(u,y,z) = u+y$, which is usually written as
$X\gets U+Y$.%
\footnote{Again, the fact that $X$ is assigned  $U+Y$ (i.e., the value
of $X$ is the sum of the values of $U$ and $Y$) does not imply
that $Y$ is assigned $X-U$; that is, $F_Y(U,X,Z) = X-U$ does not
necessarily hold.}  Thus, if $Y = 3$ and $U = 2$, then
$X=5$, regardless of how $Z$ is set.  

In the running forest fire example, suppose that we have an exogenous
random $U$ that d
etermines the values of $L$ and $\ML$.  Thus, $U$ has
four possible values of the form $(i,j)$, where both of $i$ and $j$ are
either 0 or 1.  The $i$ value determines the value of $L$ and the $j$
value determines the value of $\ML$.  Although $F_L$ gets as araguments
the vale of $U$, $\ML$, and $\FF$, in fact, it depends only
on the (first component of) the value of $U$; that is,
$F_L((i,j),m,f) = i$.  Similarly, $F_{\ML}((i,j),l,f) = j$.
%variable $U$ that determines the values 
%we would have $F_{\FF}(L,\ML,U) = 1$ if $U = (i,j,1)$ and 
%$\max(L,\ML) = 1$ (so that at least one of $L$ or $\ML$ is 1);
%similarly,  $F_{\FF}(L,\ML,U) = 1$ if $U = (i,j,2)$ and 
%$\min(L,\ML) = 1$ (so that both $L$ and $\ML$ are 1).
%Note that the
The value of $\FF$ depends only on the value of $L$ and $\ML$.
\emph{How} it depends on them depends on whether having either lightning
or an arsonist suffices for the forest fire, or whether both are
necessary.  If either one suffices, then $F_{\FF}((i,j),l,m) =
\max(l,m)$,
or, perhaps more comprehensibly, $\FF =\max(L,\ML)$; if both are needed,
then $\FF = \min(L,\ML)$.  For future reference, call the former model
the \emph{disjunctive} model, and the latter the \emph{conjunctive} model.
%value of $F_{\FF}$ is independent of the first two components of $U$;
%all that matters is the last component of $U$ and the values of $M$ and
%$\ML$.  The first two components of $U$ are used in the equations for $L$
%and $\ML$: the first determines the value of $L$ and the second determines
%the value of $\ML$, so that $F_L(\ML,\FF,(i,j,k)) = i$ and 
%$F_{\ML}(L,\FF,(i,j,k)) = j$.  Note that the values of $L$ and $\ML$ are
%independent of the value of $\FF$.  
%Because $F_X$ is a function, there is a unique value of $X$ once we have
%set all the other variables.  Notice that we have such functions only
%for the endogenous variables. The exogenous variables
%are taken as given; it is their effect on the endogenous
%variables (and the effect of the endogenous variables on each other)
%that we are modeling with the structural equations.
%}
%\end{fullv}

\fullv{
The key role of the structural equations is to define what happens in the
presence of external interventions.  For example, we can explain what
happens if the arsonist does \emph{not} drop the match.
In the disjunctive model, there is a forest fire exactly 
exactly if there is lightning; in the conjunctive model, there is
definitely no fire.
%lightning by itself suffices for a fire; that is, if the exogenous
%variable $U$ has the form $(i,1,1)$.  
Setting the value of some variable $X$ to $x$ in a causal
model $M = (\S,\F)$ results in a new causal model denoted $M_{X
\gets x}$.  In the new causal model, since the value of $X$ is
set, $X$ is removed from the list of endogenous variables.  That means
that there is no longer an equation $F_X$ defining $X$.  Moreover, $X$
is no longer an argument in the equation $F_Y$ characterizing another
endogenous variable $Y$.  The new equation for $Y$ is the one that
results by substituting $x$ for $X$.  
\shortv{See the full paper 
(available at
http://www.cs.cornell.edu/home/halpern/papers/causality+defaults.pdf) 
or \cite{HP01b} for a formal definition of 
$M_{X \gets x}$.}
More formally, 
$M_{X \gets x} = (\S_{X},
\F^{X \gets x})$, where $\S_X = (\U, \V - \{X\},  \R|_{\V - \{X\}})$
(this notation just says that $X$ is removed from the set of endogenous
variables and $\R$ is restricted so that its domain is $\V - \{X\}$
rather than all of $\V$) and  $\F^{X \gets x}$ associates with each
variable $Y \in \V- \{X\}$ the equation $F_Y^{X \gets x}$ which is 
obtained from $F_Y$ by setting $X$ to $x$.
Thus, if $M$ is the disjunctive causal model for the forest-fire example, then
$M_{\ML \gets 0}$, the model where the arsonist does not drop the match,
has endogenous variables $L$ and $\FF$, where the equation for $L$ is
just as in $M$, and $\FF \gets L$.  If $M$ is the conjunctive model,
then equation for $\FF$ becomes instead $\FF \gets 0$.

}

\journal{
\emph{Counterfactuals} (statements counter to fact) have often been
considered when trying to define causality.  These (causal) equations
can be given a straightforward counterfactual interpretation. 
An equation such as $x = F_X(\vec{u},y)$ should be thought of as saying 
that in a context where the exogenous variables have values $\vec{u}$,
if $Y$ were set to $y$ by some means (not specified in the model), 
then $X$ would take on the value $x$, as dictated by $F_X$.%
\footnote{Note that $\vec{u}$ is being used here to denote a vector, or
sequence, of values.  If there are three exogenous variables, say
$U_1$, $U_2$, and $U_3$, and $U_1 = 0$, $U_2 = 0$, and $U_3 = 1$, then
$\vec{u}$ is $(0,0,1)$.} 
However, if the value of $X$ is set by some other means (e.g., the
forest catches fire due to a volcano eruption), then 
the assignment specified by $F_X$ is ``overruled''; $Y$ is
no longer committed to tracking $X$ according to $F_X$.
This emphasizes the point that variables on the left-hand side of equations
are treated differently from ones on the right-hand side.

In the philosophy community, counterfactuals are typically defined in
terms of ``closest worlds'' \cite{Lewis73,Stalnaker68}; a statement of
the form ``if $A$ where the case then $B$ would be true'' is taken to be
true if in the ``closest world(s)'' to the actual world where $A$ is true,
$B$ is also true.  This modification of equations
may be given a simple ``closest world'' interpretation:
the solution of the equations obtained by
replacing the equation for $Y$ with the equation $Y = y$,
while leaving all other equations unaltered,
gives the closest ``world'' to the actual world where $Y=y$.
The asymmetry embodied in the structural equations can be understood in
terms of closest worlds.  If either the match or lightning suffice to
start the fire, then in the closest world to the actual world
where a lit match is dropped the forest burns down.  However, it is not
necessarily the case that in the world closest to one where the forest
burns down  a lit match is dropped.

It may seem somewhat circular to use
causal models, which clearly already encode causal relationships, to
define causality.  Nevertheless, as we shall see, there is no
circularity. In the examples typically discussed in the literature,
there is general agreement as to the appropriate causal model.  
The causal models encode background knowledge
about the tendency of certain event types to cause  other
event types (such as the fact that lightning can cause forest fires).
The models can then be used to determine the causes
of single events, such as whether it was arson
that caused the fire of June 10, 2000, given what is known or
assumed about that particular fire.%
\footnote{In general, there may be uncertainty about the causal model,
as well as about the true setting of the exogenous variables in a causal
model.  Thus, we may be uncertain about whether smoking causes cancer
(this represents uncertainty about the causal model) and uncertain about
whether a particular patient actually smoked (this is uncertainty about
the value of the exogenous variable that determines whether the patient
smokes).  This uncertainty can be described by putting a probability on
causal models and on the values of the exogenous variables.  We can then
talk about the probability that $A$ is a cause of $B$.  See also
Section~\ref{sec:blame}.}
}
                                                      
\journal{
In a causal model, it is possible that the value of $X$ can depend on
the value of $Y$ (that is, the equation $F_X$ is such that changes in
$Y$ can change the value of $X$) and the value of $Y$ can depend on the
value of $X$.  Intuitively, this says that $X$ can potentially affect
$Y$ and that $Y$ can potentially affect $X$.  While allowed by the
framework, this type of situation does not happen in the examples of
interest; dealing with it would complicate the exposition.  Thus, for
ease of exposition, I restrict attention here to what are called {\em
recursive\/} (or {\em acyclic\/}) models.  This is the special case
where there is some total ordering $\prec$ of the endogenous variables
(the ones in $\V$) 
such that if $X \prec Y$, then $X$ is independent of $Y$, 
that is, $F_X(\ldots, y, \ldots) = F_X(\ldots, y', \ldots)$ for all $y, y' \in
\R(Y)$.  Intuitively, if a theory is recursive, there is no
feedback.  If $X \prec Y$, then the value of $X$ may affect the value of
$Y$, but the value of $Y$ cannot affect the value of $X$.%
\footnote{This standard restriction is also made by HP except in the
appendix of \cite{HP01b}.}

It should be clear that if $M$ is an acyclic  causal model,
then given a \emph{context}, that is, a setting $\vec{u}$ for the
exogenous variables in $\U$, there is a unique solution for all the
equations.  We simply solve for the variables in the order given by
$\prec$. The value of the variables that come first in the order, that
is, the variables $X$ such that there is no variable $Y$ such that $
Y\prec X$, depend only on the exogenous variables, so their value is
immediately determined by the values of the exogenous variables.  
The values of values later in the order can be determined once we have
determined the values of all the variables earlier in the order.
}

In this paper, following HP, I restrict to \emph{acyclic} causal
models, where causal influence can be represented by an acyclic Bayesian
network.  That is, there is no cycle $X_1, \ldots, X_n, X_1$ of
endogenous variables where the value of $X_{i+1}$ (as given by 
$F_{X_{i+1}}$) depends on the value of $X_i$, for $1 = 1, \ldots, n-1$,
and the value of $X_1$ depends on the value of $X_n$.  If $M$ is an
acyclic  causal model, then given a \emph{context}, that is, a setting
$\vec{u}$ for the exogenous variables in $\U$, there is a unique
solution for all the equations.

\commentout{
We can describe (some salient features of) a causal model $M$ using a
{\em causal network}.%
\footnote{These causal networks are similar in spirit to the Bayesian
networks used to represent and reason about dependences in probability
distributions \cite{Pearl}.}
Figure~\ref{fig1-new} describes the causal network for
the forest fire example.   The fact that there is an edge from $U$ to
both $L$ and $\ML$ says that the value of the exogenous variable $U$
affects the value of $L$ and $\ML$, but nothing else affects it.  The
arrows from $L$ and $\ML$ to $\FF$ say that the only the values of $\ML$
and $L$ affect the value of $\FF$.  The causal network for an acyclic
causal model is \emph{acyclic}; it has no cycles.  That is, there is no
sequence of arrows that both starts and ends at the same node. 
\begin{figure}[htb]
\input{psfig}
\centerline{\includegraphics{fig1-new}}
%\begin{verbatim}
%
%   U
% /  \
%L    ML
% \  /
%  F
\caption{A simple causal network.}
\label{fig1-new}
\end{figure}
}

\fullv{
There are many nontrivial decisions to be made when choosing
the structural model to describe a given situation.
One significant decision is the set of variables used.
As we shall see, the events that can be causes and those that can be
caused are expressed in terms of these variables, as are all the
intermediate events.  The choice of variables essentially determines the
``language'' of the discussion; new events cannot be created on the
fly, so to speak.  In our running example, the fact that there is no
variable for unattended campfires means that the model does not allow us
to consider unattended campfires as a cause of the forest fire.

Once the set of variables is chosen, the next step is to decide which are
exogenous and which are endogenous.  As I said earlier, 
the exogenous variables to some extent encode the background situation
that we want to take for granted.  Other implicit background
assumptions are encoded in the structural equations themselves.
Suppose that we are trying to decide whether a lightning bolt or a
match was the
cause of the forest fire, and we want to take for granted that there is
sufficient oxygen in the air and the wood is dry.
We could model the dryness of the wood by an
exogenous variable $D$ with values $0$ (the wood is wet) and 1 (the wood
is dry).%
\footnote{Of course, in practice, we may want to allow $D$ to have more
values, indicating the degree of dryness of the wood, but that level of
complexity is unnecessary for the points I am trying to make here.}
By making $D$ exogenous, its value is assumed to be given and out of the
control of the modeler.
We could also take the amount of oxygen as an exogenous variable
(for example, there could be a variable $O$ with two values---0, for
insufficient oxygen, and 1, for sufficient oxygen); alternatively, we
could choose not to model oxygen explicitly at all.  For example,
suppose that we have, as
before, a random variable $\ML$ for match lit,
and another variable $\WB$ for wood burning,
with values 0 (it's not) and 1 (it is).  The structural equation
%$f_{\WB}$ would describe the dependence of $\WB$ on $D$
$F_{\WB}$ would describe the dependence of $\WB$ on $D$
and $\ML$.  By setting $F_{\WB}(1,1) = 1$, we are saying
that the wood will burn if the match is lit and the wood is dry.  Thus,
the equation is
implicitly modeling our assumption that there is sufficient oxygen for
the wood to burn.

According to the definition of causality in 
Section~\ref{sec:actcaus}, only endogenous variables
can be causes or be caused.  Thus, if no variables
encode the presence of oxygen, or if it is encoded only in an exogenous
variable, then oxygen cannot be a cause of the forest burning.
If we were to explicitly model the amount of oxygen in the air (which
certainly might be relevant if we were analyzing fires on Mount
Everest), then $F_{\WB}$ would also take values of $O$ as an argument,
and the presence of sufficient oxygen might well be a cause of the wood
burning, and hence the forest burning.
%\footnote{Of course, $F_{\WB}$ might take yet other variables as
%arguments.}

%Besides encoding some of our implicit assumptions, the
%structural equations can be viewed as encoding the causal mechanisms at
%work.  Changing the underlying causal mechanism can affect what counts
%as a cause. 
%Section~\ref{sec:examples} provides several
%examples of the importance of the choice of random variables and the
%choice of causal mechanism.

It is not always straightforward to decide what
the ``right'' causal model is in a given situation, nor is it 
always obvious which of two causal models is ``better'' in some sense.
These decisions  often lie at the heart of determining
actual causality in the real world.  Disagreements about causality
relationships often boil down to disagreements about the causal model.
While the formalism presented here does not provide techniques to settle
disputes about which causal model is the right one, at least it provides
tools for carefully describing the differences between causal models, 
so that it should lead to more informed and principled decisions
about those choices.

\section{A Formal Definition of Actual Cause}\label{sec:actcaus}

\subsection{A language for describing causes}
}

%\paragraph{The formal definition:}
\fullv{To make the definition of actual
causality precise, it is helpful to have a formal language for making
statements about causality.  }
%The language is a slight extension of
%\emph{propositinal logic}, now often taught in highschool.  
Given a signature $\S = (\U,\V,\R)$, a \emph{primitive event} is a
formula of the form $X = x$, for  $X \in \V$ and $x \in \R(X)$.  
%That is,
%$X$ is an endogenous variable, and $x$ is a possible value of $X$.
%The primitive event $\ML=0$ says ``the lit match is not dropped''; the
%primitive event $L=1$ says ``lightning occurred''.  As in propositional
%logic, the symbols $\land$, $\lor$, and $\neg$ are used to denote
%conjunction, disjunction, and negation, respectively.  Thus, the formula
%$\ML=0 \lor L=1$ says ``either the lit match is not dropped or lightning
%occurred'', $\ML=0 \land L=1$ says ``either the lit match is not dropped
%and lightning occurred'', and $\neg (L=1)$ says ``lightning did not
%occur'' (which is equivalent to $L=0$ given that the only possible
%values of $L$ are 0 or 1).  A \emph{Boolean combination} of primitive
%events is a formula 
%that is obtained by combining primitive events using $\land$, $\lor$,
%and $\neg$.  Thus, $\neg(\ML=0 \lor L=1) \land \WB=1$ is Boolean
%combination of the primitive events $\ML=0$, $L=1$, and $\WB=1$.
%
A {\em causal formula (over $\S$)\/} is one of the form
$[Y_1 \gets y_1, \ldots, Y_k \gets y_k] \phi$,
where
\fullv{
\begin{itemize}
\item}
$\phi$ is a Boolean
combination of primitive events,
\fullv{\item} $Y_1, \ldots, Y_k$ are distinct variables in $\V$, and
\fullv{\item} $y_i \in \R(Y_i)$.
\fullv{\end{itemize}}
%and $\vec{u}$ is a vector of values for all the variables in $\U$.
Such a formula is
abbreviated
%as $[\vec{Y} \gets \vec{y}]\phi(\vec{u})$.
as $[\vec{Y} \gets \vec{y}]\phi$.
The special
case where $k=0$
is abbreviated as
%$[\true]\phi(\vec{u})$. Intuitively, $[\true]\phi(\vec{u})$ says that
$\phi$.
Intuitively,
$[Y_1 \gets y_1, \ldots, Y_k \gets y_k] \phi$ says that
%$\phi(\vec{u})$ holds in the counterfactual world that would arise if
$\phi$ would hold if
$Y_i$ were set to $y_i$, for $i = 1,\ldots,k$.

A causal formula $\psi$ is true or false in a causal model, given a
context.
As usual, I write $(M,\vec{u}) \sat \psi$  if
the causal formula $\psi$ is true in
causal model $M$ given context $\vec{u}$.
The $\sat$ relation is defined inductively.
%\fullv{
%Perhaps not surprisingly, $(M,\vec{u}) \sat \psi$ is read ``$\psi$ is
%true in context $\vec{u}$ in causal model model $M$''.}
%\end{fullv}
$(M,\vec{u}) \sat X = x$ if
the variable $X$ has value $x$
in the
unique (since we are dealing with acyclic models) solution
to
the equations in
$M$ in context $\vec{u}$
%\fullv{$\vec{u}$ 
(that is, the
unique vector
of values for the exogenous variables that simultaneously satisfies all
equations 
in $M$ 
with the variables in $\U$ set to $\vec{u}$).
The truth of conjunctions and negations is defined in the standard way.
Finally, 
%\fullv{$(M,\vec{u}) \sat
%[\vec{Y} \gets \vec{y}]\phi$ for an arbitrary Boolean combination
%$\phi$ of formulas of the form $\vec{X} = \vec{x}$ is defined
%similarly.}
$(M,\vec{u}) \sat [\vec{Y} \gets \vec{y}]\phi$ if 
$(M_{\vec{Y} \gets \vec{y}},\vec{u}) \sat \phi$.
%combination of primitive
%event, is defined similarly.  The truth of Boolean combinations
%of causal formulas is defined in the obvious way.
%\end{fullv}  
I write $M \sat \phi$ if $(M,\vec{u}) \sat \phi$ for all contexts $\vec{u}$.

\fullv{
For example, if $M$ is the disjunctive causal model for the forest
fire, and $u$ is the context where there is lightning and the arsonist
drops the lit match, 
then $(M,u) \sat [\ML \gets 0](\FF=1)$, since even if the arsonist is somehow
prevented from dropping the match, the forest burns (thanks to the
lightning);  similarly, $(M,u) \sat [L \gets 0](\FF=1)$.  However,
$(M,u) \sat [L \gets 0;\ML \gets 0](\FF=0)$: if arsonist does not drop
the lit match 
and the lightning does not strike, then the forest does not burn.

\subsection{A preliminary definition of causality}
The HP definition of causality, like many others, is based on
counterfactuals.  The idea is that $A$ is a cause of $B$ if, if $A$
hadn't occurred (although it did), then $B$ would not have occurred.
This idea goes back to at least Hume \citeyear[Section
{VIII}]{hume:1748}, who said:
\begin{quote}
We may define a cause to  be an object followed
by another, \ldots, if the first object had not been, the second
never had existed.
\end{quote}
This is essentially the \emph{but-for} test, perhaps the most widely
used test of actual causation in tort adjudication.  The but-for test
states that an act is a cause of injury if and only if, but for the act
(i.e., had the the act not occurred), the injury would not have
occurred.

There are two well-known problems with this definition.  The first can
be seen by considering the disjunctive causal model for the forest fire
again.  Suppose that the 
arsonist drops a match and lightning strikes.
Which is the cause?
According to a naive interpretation of the counterfactual definition,
neither is.  If the match hadn't dropped, then the lightning would
still have struck, so there would have been a forest fire anyway.
Similarly, if the lightning had not occurred, there still would have
been a forest fire.  As we shall see, the HP definition declares
both lightning and the 
arsonist cases of the fire.  (In general, there may be more than one
cause of an outcome.)

A more subtle problem is what philosophers have called
\emph{preemption}, where there are two potential causes
of an event, one of which preempts the other.  Preemption 
is illustrated by the following story taken from \cite{Hall98}:

\begin{quote}
Suzy and Billy both pick up rocks
and throw them at  a bottle.
Suzy's rock gets there first, shattering the
bottle.  Since both throws are perfectly accurate, Billy's would have
shattered the bottle had it not been preempted by Suzy's throw.
\end{quote}
Common sense suggests that Suzy's throw is the cause of the shattering,
but Billy's is not.  However, it does not satisfy the naive counterfactual
definition either; if Suzy hadn't thrown, then Billy's throw would have
shattered the bottle.

The HP definition deals with the first  problem by defining causality as 
counterfactual dependency \emph{under certain contingencies}.
In the forest fire example, the forest fire does counterfactually depend
on the lightning under the contingency that the arsonist does not drop
the match; similarly, the forest fire depends oounterfactually on the
arsonist's match under the contingency that the lightning does not
strike.  Clearly we need to be a little careful here to limit the
contingencies that can be considered.  We do not want to make Billy's
throw the cause of the bottle shattering by considering the contingency
that Suzy does not throw.   The reason that we consider Suzy's throw to
be the cause and Billy's throw not to be the cause is that Suzy's rock
hit the bottle, while Billy's did not.  Somehow the definition must
capture this obvious intuition.

With this background, I now give the preliminary version of the HP definition
of causality.  Although the definition is labeled ``preliminary'',
it is quite close to the final definition, which is given 
in Section~\ref{sec:final}.  As I pointed out in the
introduction, the definition is relative to a causal model (and a
context); $A$ may be a cause  of $B$ in one causal model but not in another.
The definition consists of three clauses.  The first and third are quite
simple; all the work is going on in the second clause.  

The types of events that the HP definition allows as actual causes are
ones of the form $X_1 = x_1 \land \ldots \land X_k = x_k$---that is,
conjunctions of primitive events; this is often abbreviated as $\vec{X}
= \vec{x}$. The events that can be caused are arbitrary Boolean
combinations of primitive events.
The definition does not allow statements of the form  ``$A$ or $A'$ is a
cause of $B$,'' although this could be treated as being equivalent to
``either $A$ is a cause of $B$ or $A'$ is a cause of $B$''.    
%Essentially, this is interpreted as ``either $A$ is a cause of $B$ or
%$A'$ is a cause of $B$'', and then evaluate the truth of ''$A$ is a cause
%of $B$'' and ``$A'$ is a cause of $B$'' separately.  
On the other hand, statements such as
``$A$ is a cause of $B$ or $B'$'' are allowed;  as we shall see, this is not
equivalent to ``either $A$ is a cause of $B$ or $A$ is a cause of $B'$''.
}
%\end{fullv}

\dfn\label{actcaus}
(Actual cause; preliminary version) \cite{HP01b}
$\vec{X} = \vec{x}$ is an {\em actual cause of $\phi$ in
$(M, \vec{u})$ \/} if the following
three conditions hold:
\begin{description}
\item[{\rm AC1.}]\label{ac1} $(M,\vec{u}) \sat (\vec{X} = \vec{x})$ and 
$(M,\vec{u}) \sat \phi$.
%(That is, for $\vec{X} = \vec{x}$ to be a cause of $\phi$, both $\vec{X}
%= \vec{x}$ and $\phi$ are true in the actual context.)
\item[{\rm AC2.}]\label{ac2}
There is a partition of $\V$ (the set of endogenous variables) into two
%There exist disjoint sets $\vec{Z}$ and $\vec{W}$ of endogenous variables
subsets $\vec{Z}$ and $\vec{W}$  
with $\vec{X} \subseteq \vec{Z}$ and a
setting $\vec{x}'$ and $\vec{w}$ of the variables in $\vec{X}$ and
$\vec{W}$, respectively, such that
if $(M,\vec{u}) \sat Z = z^*$ for 
all $Z \in \vec{Z}$, then
both of the following conditions hold:
\begin{description}
\item[{\rm (a)}]
$(M,\vec{u}) \sat [\vec{X} \gets \vec{x}',
\vec{W} \gets \vec{w}]\neg \phi$.
%In words, changing $(\vec{X},\vec{W})$ from $(\vec{x},\vec{w})$ to
%$(\vec{x}',\vec{w}')$ changes
%$\phi$ from true to false.
\item[{\rm (b)}]
$(M,\vec{u}) \sat [\vec{X} \gets
\vec{x}, \vec{W}' \gets \vec{w}, \vec{Z}' \gets \vec{z}^*]\phi$ for 
all subsets $\vec{W}'$ of $\vec{W}$ and all subsets $\vec{Z'}$ of
$\vec{Z}$, where I abuse notation and write $\vec{W}' \gets \vec{w}$ to
denote the assignment where the variables in $\vec{W}'$ get the same
values as they would in the assignment $\vec{W} \gets \vec{w}$. 
\end{description}
\item[{\rm AC3.}] \label{ac3}
$\vec{X}$ is minimal; no subset of $\vec{X}$ satisfies
conditions AC1 and AC2.
\label{def3.1}  %%changed from {def2.2}
\end{description}
$\vec{W}$, $\vec{w}$, and $\vec{x}'$ are said to be
\emph{witnesses} to the fact that $\vec{X} = \vec{x}$ is a cause of
$\phi$.
\end{definition}

AC1 just says that $\vec{X}=\vec{x}$ cannot
be considered a cause of $\phi$ unless both $\vec{X} = \vec{x}$ and
$\phi$ actually happen.  AC3 is a minimality condition, which ensures
that only those elements of 
the conjunction $\vec{X}=\vec{x}$ that are essential for
changing $\phi$ in AC2(a) are
\shortv{considered part of a cause.}
\journal{considered part of a cause; inessential elements are pruned.
Without AC3, if dropping a lit match qualified as a
cause of the forest fire, then dropping a match and
sneezing would also pass the tests of AC1 and AC2.
AC3 serves here to strip ``sneezing''
and other irrelevant, over-specific details
from the cause.}  
\fullv{Clearly, all the ``action'' in the definition occurs in AC2.
We can think of the variables in $\vec{Z}$ as making up the ``causal
path'' from $\vec{X}$ to $\phi$.  Intuitively, changing the value of
some variable in $X$ results in changing the value(s) of some
variable(s) in $\vec{Z}$, which results in the values of some
other variable(s) in $\vec{Z}$ being changed, which finally results in
the value of $\phi$ changing.  The remaining endogenous variables, the
ones in $\vec{W}$, are off to the side, so to speak, but may still have
an indirect effect on what happens.}
AC2(a) is essentially the standard
counterfactual definition of causality, but with a twist.  If we 
want to show that $\vec{X} = \vec{x}$ is a cause of $\phi$, we must show
(in part) that if $\vec{X}$ had a different value, then so too would
$\phi$.  However, this effect of the value of $\vec{X}$ on the value of
$\phi$ may not hold in the actual context;
\fullv{the value of $\vec{W}$ may have to be different to allow this
effect to manifest itself.  For example, consider 
the context where both the lightning
strikes and the arsonist drops a match in the disjunctive model of the
forest fire.  Stopping the arsonist from
dropping the match will not prevent the forest fire.  The
counterfactual effect of the arsonist on the forest fire manifests
itself only in a situation where the lightning does not strike (i.e., where
$L$ is set to 0).  AC2(a) is what allows us to call both the
lightning and the arsonist causes of the forest fire.
}
%\shortv{AC2(a) allows us to show the counterfactual dependence of $\phi$
%on $\vec{X} = \vec{x}$ under the contingency $\vec{W} = \vec{w}$.
%AC2(b) is perhaps the most complicated condition.  It limits the
%``permissiveness'' of AC2(a) with regard to the 
%contingencies that can be considered.}
Essentially, it ensures that
$\vec{X}$ alone suffices to bring about the change from $\phi$ to $\neg
\phi$; setting $\vec{W}$ to $\vec{w}$ merely eliminates
possibly spurious side effects that may mask the effect of changing the
value of $\vec{X}$.  Moreover, although the values of variables on the
causal path (i.e.,  the variables $\vec{Z}$) may be perturbed by
the change to $\vec{W}$, this perturbation has no impact on the value of
$\phi$.  If  $(M,\vec{u}) \sat \vec{Z} = \vec{z}^*$, then $\vec{z}^*$
is the value of the 
variable $Z$ in the context $\vec{u}$.  We capture the fact that the
perturbation has no impact on the value of $\phi$ by saying that if some
variables $Z$ on the causal path were set to their original values in the
context $\vec{u}$, $\phi$ would still be true, as long as $\vec{X} =
\vec{x}$.
\shortv{Roughly speaking, it says that if the variables in $\vec{X}$ are reset
to their original value, then $\phi$ holds, even under the contingency
$\vec{W}' = \vec{w}$ and even if some variables in $\vec{Z}$ are given
their original values (i.e., the values in $\vec{z}^*$).}

\journal{
This condition is perhaps best understood by considering the
Suzy-Billy example.  It is an attempt to capture the intuition that the
reason that Suzy is the cause of the bottle shattering, and not Billy,
is that Suzy's rock actually hit the bottle and Billy's didn't.  
If there is a variable in the model that represents whether Billy's rock
hits the bottle, then for Billy's throw to be a cause of the bottle
shattering, the bottle would have to shatter (that will be the $\phi$ in
AC2(b)) if Billy throws ($X=x$) and Suzy does not ($W = w$) even if, as
is actually the case in the real world, Billy's rock does not hit the
bottle ($Z = z^*$).  
%This should become clearer when we analyze this
%example more formally, in Example~\ref{xam2}.   Before considering this
%examples, I start with a simpler example, just to give the intuitions.  
}

\journal{
\xam\label{ex:ff}
For the forest fire example, 
let  $M$ be the causal model for the forest fire sketched earlier, 
with endogenous variables $L$, $\ML$, and $\FF$.
Clearly $(M,(1,1,1)) \sat \FF=1$ and 
$(M,(1,1,1)) \sat L=1$; in the context (1,1,1), the lightning strikes
and the forest burns down.  Thus, AC1 is satisfied.  AC3 is trivially
satisfied, since $\vec{X}$ consists of only one element, $L$, so must be
minimal.  For AC2, take $\vec{Z} = \{L, \FF\}$ and take $\vec{W} =
\{\ML\}$, let $x' = 0$, and let $w = 0$.  Clearly,
$(M,(1,1,1)) \sat [L \gets 0, \ML \gets 0](\FF \ne 1)$; if the lightning
does not strike and the match is not dropped, the forest does not burn
down, so AC2(a) is satisfied.  To see the effect of the lightning, we
must consider the contingency where the match is not dropped; the
definition allows us to do that by setting $\ML$ to 0.  (Note that here
setting $L$ and $\ML$ to 0 overrides the effects of $U$; this is
critical.)  Moreover,  
$(M,(1,1,1)) \sat [L \gets 1, \ML \gets 0](\FF = 1)$;
if the lightning
strikes, then the forest burns down even if the lit match is not
dropped, so AC2(b) is satisfied.  (Note that since $\vec{Z} = \{L, \FF\}$, 
the only subsets of $\vec{Z} - \vec{X}$ are the empty set and the
singleton set consisting of just $\FF$.)

The lightning and the dropped match are also causes of the forest fire
in the context where $U = (1,1,2)$, where both the lightning and 
match are needed to start the fire.  
Again, I just present the argument for the lightning here.
And, again, both AC1 and AC3 are trivially satisfied.  For AC2, again
take $\vec{Z} = \{L, \FF\}$, $\vec{W} =
\{\ML\}$, and $x' = 0$, but now let  $w = 1$.
We have that 
$$(M,(1,1,2)) \sat [L \gets 0, \ML \gets 1](\FF \ne 1) \mbox{ and }$$
$$(M,(1,1,2)) \sat [L \gets 1, \ML \gets 1](\FF = 1),$$
so AC2(a) and AC2(b) are satisfied.  

As this example shows, causes are not
unique; there may be more than one cause of a given outcome.  Moreover,
both the lightning and the dropped match are causes both 
in the case where either one suffices to start the fire and in the
case where both are needed.  As we shall see, the notion of
\emph{responsibility} distinguishes these two situations.  Finally, it
is worth noting that the lightning is not the cause in either the context
$(1,0,2)$ or the context $(1,1,0)$.  In the first case, AC1 is violated.
If both the lightning and the match are needed to cause the fire, then
there is no fire if the match is not dropped.   In the second case,
there is a fire but, intuitively, it arises spontaneously; neither the
lightning nor the dropped match are needed.  Here AC2(a) is violated;
there is no setting of $L$ and $\ML$ that will result in no forest fire.
\exam

\xam\label{xam2}
Now let us consider the Suzy-Billy example.%
\footnote{The discussion of this and the following example is taken
almost verbatim from HP.}
We get the desired result---that Suzy's throw is a cause, but Billy's is
not---but only if we 
model the story appropriately.  Consider first a coarse causal
model, with three endogenous variables:
\begin{itemize}
\item $\ST$ for ``Suzy throws'', with values 0 (Suzy does not throw) and
1 (she does);
\item $\BT$ for ``Billy throws'', with values 0 (he doesn't) and
1 (he does);
\item $\BS$ for ``bottle shatters', with values 0 (it doesn't shatter)
and 1 (it does).
\end{itemize}
(I omit the exogenous variable here; it determines whether Billy and
Suzy throw.)  Take the formula for $\BS$ to be such that the bottle
shatters if either Billy or Suzy throw; that is $\BS = \BT \vee \ST$. (I
am implicitly assuming that Suzy and Billy never miss if they throw.)
$\BT$ and $\ST$ play symmetric roles in this model;
there is nothing to distinguish them.
Not surprisingly, both
Billy's throw and Suzy's throw are classified as causes of the
bottle shattering
in this model.  The argument is essentially identical to the forest fire
example in the case that $U = (1,1,1)$, where either the lightning or
the dropped match is enough to start the fire.

The trouble with this model is that it cannot distinguish
the case where both rocks
hit the bottle simultaneously (in which case it would be reasonable
to say that both $\ST=1$ and $\BT=1$ are 
causes of $\BS=1$) from the case where Suzy's rock
hits first.  The model has to be refined to express this distinction.
One way is to invoke a dynamic model \cite[p.~326]{pearl:2k}.  This
model is discussed by HP; I omit details here. 
A perhaps simpler way to gain expressiveness is to allow $\BS$ to be
three valued, with values 0 (the bottle doesn't shatter), 1
(it shatters as a result of being hit by Suzy's rock), and 2 (it
shatters as a result of being hit by Billy's rock).
I leave it to the reader to check that $\ST = 1$ is a
cause of $\BS = 1$, but $\BT = 1$ is not (if Suzy doesn't thrown but
Billy does, then we would have $\BS = 2$).  Thus, to some extent, this
solves our problem.  But it
borders on cheating; the answer is almost
programmed into the model by invoking the relation ``as a result of'',
which requires the identification of the actual cause.

A more useful choice is to add two new random variables to the model:
\begin{itemize}
\item $\BH$ for ``Billy's rock hits the (intact) bottle'', with values 0
(it doesn't) and 1 (it does); and
\item $\SH$ for ``Suzy's rock hits the bottle'', again with values 0 and
1.
\end{itemize}
Now it is the case that, in the context where both Billy and Suzy throw,
$\ST=1$ is a cause 
of $\BS=1$, but $\BT = 1$ is not.
To see that $\ST=1$ is a cause, note that, as usual, it is immediate
that AC1 and AC3 hold.  For AC2, choose $\vec{Z} = \{\ST,\SH,\BH\}$,
$\vec{W}=\{\BT\}$,  and $w=0$. 
When $\BT$ is set to 0, $\BS$ tracks $\ST$: if Suzy
throws, the bottle shatters  and if she doesn't throw, the bottle does
not shatter.  To see that $\BT=1$ is \emph{not} a cause
of $\BS=1$, we must check that there is no
partition $\vec{Z} \cup \vec{W}$ of the endogenous variables that
satisfies AC2.
Attempting the symmetric choice with $\vec{Z} = \{\BT,\BH,\SH\}$,
$\vec{W}=\{\ST\}$, and $w=0$ 
violates AC2(b). To see this, take $\vec{Z}' = \{\BH\}$.  In the
context where Suzy 
and Billy both throw, $\BH=0$.  If $\BH$ is set to 0, the bottle does
not shatter if Billy throws and Suzy does not.  
It is precisely because, in this context, Suzy's throw hits the bottle
and Billy's does not that we declare Suzy's throw to be the cause of the
bottle shattering.  AC2(b) captures that intuition by allowing us to
consider the contingency where $\BH=0$, despite the fact that Billy
throws.   I leave it to the reader to check that no other partition of
the endogenous variables satisfies AC2 either.  

%This example illustrates the need for requiring in AC2(b) that $\phi$
%continue to hold even if subsets of $\vec{Z}$ are set to their original
%value.  Let $\vec{Z} = \{\BT,\BH,\BS\}$, $\vec{W} = \{\ST,\SH\}$, and
%$\vec{w} = (0,0)$ in the Suzy-Billy
%$(M,\vec{u}) \sat [\vec{X} \gets \vec{x}, \vec{W} \gets \vec{w}]\phi$
%holds if we take 
%and thus without the requirement that AC2(b) hold for all subsets of
%$\vec{Z}$, $\BT=1$ would have qualified as a cause of $\BS=1$.
%Insisting that $\phi$ remains unchanged
%when both $\vec{W}$ is set to
%$\vec{w}$ and $\vec{Z}'$ is set to $\vec{z}^*$ (for an arbitrary subset
%$\vec{Z}'$ of $\vec{Z}$)
%prevents us from choosing contingencies
%$\vec{W}$ that interfere with the active causal paths from $\vec{X}$ to
%$\phi$.

This example emphasizes an important moral.
If we want to argue in a case of preemption
that $X=x$ is the cause of $\phi$ rather than $Y=y$,
then there must be a random variable ($\BH$ in this case) that takes on
different values depending on whether $X=x$ or $Y=y$ is the actual
cause.  If the model does not contain such a variable, then it will not
be possible to determine which one is in fact the cause.  This is
certainly consistent with intuition and the way we present evidence.  If
we want to  argue (say, in a court of law) that it was $A$'s shot that
killed $C$ rather than $B$'s, then we present evidence such as the
bullet entering $C$ from the left side (rather than the right side, which
is how it would have entered had $B$'s shot been the lethal one).
The side from which the shot entered is the relevant random variable in
this case.  Note that the random variable may involve temporal evidence
(if $Y$'s shot had been the lethal one, the death would have occurred
a few seconds later), but it certainly does not have to.
\exam
}
%\end{fullv}

To give some intuition for this definition, I consider three examples that
will be relevant later in the paper.

\xam\label{xam4}  
\fullv{Can {\em not} performing an action be (part of) a cause?
Consider the following story, also taken from
(an early version of) \cite{Hall98}:}
Suppose that Billy is hospitalized with a mild illness on Monday; he is
treated and recovers.  In the obvious causal model, the doctor's
treatment is a cause of Billy's recovery.  Moreover, if the doctor
does \emph{not} treat Billy on Monday, then the doctor's omission to
treat Billy is a cause of Billy's being sick on Tuesday.
%It seems that it should be, and indeed it is
%according to our analysis.  Suppose that $\vec{u}$ is the context where,
%among other things, Billy is sick on Monday and the situation is such
%that the doctor forgets to administer the medication Monday.
%(There is much more to the
%context $\vec{u}$, as we shall shortly see.)  It seems reasonable that
%the model should have two random variables:
%\begin{itemize}
%\item $\MT$ for ``Monday treatment'', with values 0 (the doctor does
%not treat Billy on Monday) and 1 (he does); and
%\item $\BMC$ for ``Billy's medical condition'', with values 0 (recovered)
%and 2 (dead).  (We don't need the third value
%\end{itemize}
%Sure enough, in the obvious causal model, $\MT=0$ is a cause
%Again, in the obvious causal model, it is.
%This may seem somewhat disconcerting at first.  
But now suppose there are 100
doctors in the hospital.  Although only doctor 1 is assigned to
Billy (and he forgot to give medication), in principle, any of the other
99 doctors could have given Billy his medication.  
%Is the fact that they
%didn't give him the medication also part of the cause 
%of him still being sick on Tuesday?
Is the nontreatment by doctors 2--100 also a cause of Billy's being
sick on Tuesday?
%
%In the causal model above, the other doctors' failure to give Billy his
%medication is not a cause, since the model has no random variables to
%model the other doctors' actions, just as there was  no random variable
%in the causal model of Example~\ref{ex:ff} to model the presence of oxygen. 
%Their lack of action is part of the context.  We factor it out because
%(quite reasonably) we want to focus on the actions of Billy's doctor.
%If we had included endogenous random variables corresponding to the
%other doctors, then they too would be causes of Billy's
%being sick on Tuesday.  
Of course, if we do not have variables in the model corresponding to the
other doctors' treatment, or treat these variables as exogenous, then
there is no problem.  But if we have endogenous variables
corresponding to the other doctors (for example, if we want to also 
consider other patients, who are being treated by these other doctors),
then the other doctors' nontreatment is a cause, which seems inappropriate.
I return to this issue in the next section.
\shortv{\exam}

\fullv{
With this background, we
continue with Hall's modification of the original story.

\begin{quote}
Suppose that Monday's doctor is reliable, and administers the medicine
first thing in the morning, so that Billy is fully recovered by Tuesday
afternoon.  Tuesday's doctor is also reliable, and would have treated
Billy if Monday's doctor had failed to.  \ldots And let us add a twist:
one dose of medication is harmless, but two doses are lethal.
\end{quote}
Is the fact that Tuesday's doctor did {\em not\/}
treat Billy the cause of him being alive (and recovered) on
Wednesday morning?

The causal model for this story is straightforward.
There are three random variables: 
\begin{itemize}
\item $\MT$ for Monday's treatment (1 if
Billy was treated Monday; 0 otherwise); 
\item $\TT$ for Tuesday's treatment (1
if Billy was treated Tuesday; 0 otherwise); and 
\item $\BMC$ for Billy's
medical condition
(0 if Billy is fine both Tuesday morning and Wednesday morning;
1  if Billy is
sick Tuesday morning, fine Wednesday morning; 2 if Billy
is sick both Tuesday and Wednesday morning; 3  if Billy is fine
Tuesday morning and dead Wednesday morning).
\end{itemize}
We can then describe Billy's condition as a function of the four
possible combinations of treatment/nontreatment on Monday and Tuesday.
I omit the obvious structural equations corresponding to this
discussion.  

In this causal model, it is true that $\MT=1$ is a cause
of $\BMC=0$, as we would expect---because Billy is treated Monday, he is not
treated on Tuesday morning, and thus recovers Wednesday morning.
$\MT=1$ is also a cause
of $\TT=0$, as we would
expect, and $\TT=0$ is a cause
of Billy's being alive ($\BMC=0
\lor \BMC=1 \lor \BMC=2$).  However, $\MT=1$ is {\em not\/} a cause
of Billy's being alive.  It fails condition AC2(a): setting
$\MT=0$ still leads to Billy's being alive (with $W=\emptyset$).
Note that it would not help to take $\vec{W}= \{\TT\}$.  For if $\TT=0$,
then Billy is alive no matter what $\MT$ is, while if $\TT=1$, then Billy is
dead when $\MT$ has its original value, so AC2(b) is violated (with
$\vec{Z}' = \emptyset$).

This shows that
causality is not transitive, according to our definitions.
Although $\MT=1$ is a cause of $\TT=0$ and $\TT=0$ is a
cause of $\BMC=0 \lor \BMC=1 \lor \BMC=2$, $\MT=1$ is not a cause
of $\BMC=0 \lor \BMC=1 \lor \BMC=2$.
Nor is causality closed
under {\em right weakening}:  $\MT=1$ is a cause of $\BMC=0$,
which logically implies $\BMC=0 \lor \BMC=1 \lor \BMC=2$, which is not
caused by $\MT=1$.  

This distinguishes the HP definition from that of Lewis
\citeyear{Lewis00}, which builds in
transitivity and implicitly assumes right weakening.  
\exam
}
%\end{fullv}

The version of AC2(b) used here is taken from \cite{HP01b}, and differs
from the version given in the conference version of that paper \cite{HPearl01a}.
In the current version, AC2(b) is required to hold for all subsets
$\vec{W}'$ of $\vec{W}$; in the original definition, it was required to
hold only for $\vec{W}$.  The following example, due to Hopkins and Pearl
\citeyear{HopkinsP02}, illustrates why the change was made.

\xam\label{xam3} Suppose that a prisoner dies 
either if $A$ loads $B$'s gun and $B$ shoots, or if $C$ loads and shoots
his gun.  Taking $D$ to represent the prisoner's death and making the
obvious assumptions about the meaning of the variables, we have that
$D=1$ iff $(A=1 \land B=1) \lor (C=1)$.  Suppose that in the actual
context $u$, $A$ loads $B$'s gun, $B$ does not shoot, but $C$ does load
and shoot his gun, so that the prisoner dies.  Clearly $C=1$ is a cause
of $D=1$.  We would not want to say that $A=1$ is a cause of $D=1$ in
context $u$; given that $B$ did not shoot (i.e., given that $B=0$),
$A$'s loading the gun should not count as a cause.  The obvious way to
attempt to show that $A=1$ is a cause is to take $\vec{W} = \{B,C\}$ and
consider the contingency where $B=1$ and $C=0$.
It is easy to check that AC2(a) holds for this contingency; moreover,
$(M,u) \sat [A \gets 1, B \gets 1, C \gets 0](D=1)$.  However,
$(M,u) \sat [A \gets 1, C \gets 0](D=0)$.  Thus, AC2(b) is not satisfied
for the subset $\{C\}$ of $W$, so $A=1$ is not a cause of $D=1$.  However,
had we required AC2(b) to hold only for $\vec{W}$ rather than all
subsets $\vec{W}'$ of $\vec{W}$, then $A=1$ would have been a cause.
\exam

While the change in AC2(b) has the advantage of being able to deal with
Example~\ref{xam3} (indeed, it deals with the whole class of examples
given by Hopkins and Pearl of which this is an instance), it has a
nontrivial side effect.    
For the original definition, it was shown that the minimality condition 
AC3 guarantees that causes are always single conjuncts 
\fullv{\cite{EL01,Hopkins01}.  It was claimed in \cite{HP01b} that the
result}
\shortv{\cite{EL01}.  It was claimed in \cite{HP01b} that the result}
is still true for the modified definition, but, as I now show, this is
not the case. 

\xam\label{xam3b} $A$ and $B$ both vote for a candidate.  $B$'s vote is
recorded in two optical scanners ($C_1$ and $C_2$).  If $A$ votes for the
candidate, 
then she wins; if $B$ votes for the candidate and his vote is correctly
recorded in the optical scanners,
then the candidate wins.  
Unfortunately, $A$
also has access to the scanners, so she will set them to read 0 if she does
not vote for the candidate.  In the actual context $\vec{u}$, both $A$
and $B$ vote for the candidate.   The following structural equations
characterize $C$ and $\WIN$: $C_i=\min(A,B)$, $i = 1,2$, and $\WIN = 1$
iff $A=1$ or $C_1=C_2=1$.   
%\shortv{As I show in the full paper (available on the AAAI web site)}
I claim that
$C_1=1 \land C_2=1$ is a cause of $\WIN=1$,
but neither $C_1=1$ nor $C_2=1$ is a cause.  
%\fullv{
To see that $C_1=1 \land C_2=1$ is
a cause, first observe that AC1 clearly holds.  For AC2, let $\vec{W} =
\{A\}$ (so $\vec{Z} = \{B,C_1,C_2,\WIN\}$) and take $w = 0$ (so we are
considering 
the contingency where $A=0$).  Clearly, $(M,\vec{u}) \sat [C_1 \gets 0, C_2
\gets 0, A \gets 0](\WIN=0)$ and $(M,\vec{u}) \sat [C_1 \gets
1, C_2 \gets 1, A \gets a](\WIN=1)$, for both $a=0$ and $a=1$, so AC2
holds.  To show that AC3 
holds, I must 
show that neither $C_1=1$ nor $C_2=1$ is a cause of $\WIN=1$.  The
argument is the same for both $C_1=1$ and $C_2=1$, so I just show that 
$C_1 = 1$ 
is not a cause.  To see this, note that if $C_1=1$ is a cause with $\vec{W}$,
$\vec{w}$, and $\vec{x}'$ as witnesses, then  $\vec{W}$ must contain $A$
and $\vec{w}$ must be such that $A=0$.   But since $(M,u) \sat [C_1 \gets
1, A \gets 0](\WIN = 0)$, AC2(b) is violated no matter whether $C_2$ is in
$\vec{Z}$ 
or in $\vec{W}$. 
%Similarly, if $C=1$ is a cause with $\vec{W}$,
%$\vec{w}$, and $\vec{x}'$ as witnesses, then  $\vec{W}$ must again
%contain $B$ and $\vec{w}$ must again be such that $B=0$.   But since
%$(M,u) \sat [C \gets 1, B \gets 
%0](\WIN = 0)$, AC2(b) is violated no matter whether $A$ is in $\vec{Z}$
%or in $\vec{W}$.
\exam

Although Example~\ref{xam3b} shows that causes are not always single
conjuncts, they often are.  Indeed, it is not hard to show that in all
the standard examples considered in the philosophy and legal literature
(in particular, in all the examples considered in HP), they are.  
\shortv{This phenomenon is explained in Section~\ref{sec:NESS}.}
\fullv{
The following result give some intuition as to why.  Further intuition
is given by the results of Section~\ref{sec:NESS}.
Notice that in Example~\ref{xam3b}, $A$ affects both $C_1$ and $C_2$.  
%This turns out to be the key to getting causes that are not single
%conjuncts. 
As the following result shows, we do not have conjunctive causes if the
potential causes cannot be affected by other variables.

\commentout{
\dfn 
If $M = (\U,\V,\F)$ is a causal model and 
%$X, y \in\U \union \V$, then 
%$X$ is \emph{independent of $Y$
%% conditional on $\vec{Z} 
%in $M$} 
%if, for  all values $\vec{z}$ of the variables in $\U \union \V - \{X,Y\}$
%and all values $y$ and $y'$ of $Y$,
%we have $F_X(\vec{z},y)= F_X(\vec{z},y')$.  
%$X$ is \emph{dependent} on $Y$ if it is not independent of $Y$.
%If 
$X \union \vec{Y} \subseteq \V$, then 
$X$ \emph{is not affected by} $\vec{Y}$ in $(M,\vec{u}$ if
$(M,\vec{u}) \sat X = x$ implies $(M,\vec{U}) \sat [\vec{Y} \gets
\vec{y}](X=x)$ for 
all possible settings $\vec{y}$ of the variables in $\vec{Y}$.  That is,
setting $\vec{Y}$ to an arbitrary value does not 
cause the value of $X$ to change in context $\vec{u}$ in $M$.
\edfn
}

Say that $\vec{X}
= \vec{x}$ is a \emph{weak cause of $\phi$ under the contingency 
$\vec{W} = \vec{w}$ in $(M,\vec{u})$} if AC1 and AC2 hold under
the contingency $\vec{W} = \vec{w}$, but AC3 does not necessarily hold.

\pro\label{pro:singlecause1} If $\vec{X} = \vec{x}$ is a weak cause of
$\phi$ in $(M,\vec{u})$ with $\vec{W}$, $\vec{w}$, and $\vec{x}'$ as
witnesses, $|\vec{X}| > 1$, and each variable  $X_i$ in $\vec{X}$
is independent of all the variables in $\V - \vec{X}$ in $\vec{u}$
(that is, if $\vec{Y} \subseteq \V - \vec{X}$, then for each setting
$\vec{y}$ of $\vec{Y}$, we have
$(M,\vec{u}) \sat \vec{X} = \vec{x}$ iff $(M,\vec{u}) \sat [\vec{Y} \gets
\vec{y}] (\vec{X} = \vec{x})$), 
then $\vec{X} =
\vec{x}$ is not a cause of $\phi$ in $(M,\vec{u})$. 
\epro
 
In the examples in \cite{HP01b} (and elsewhere in the literature), the
variables that are potential causes are typically independent of
all other variables, so in these causes are in fact single conjuncts.
}

\section{Dealing with normality and typicality}\label{sec:final}

While the definition of causality given in Definition~\ref{actcaus}
works well in many cases, it does not always deliver answers that
agree with (most people's) intuition.  Consider the following example,
taken from Hitchcock \citeyear{Hitchcock07}, based on an example due
to Hiddleston \citeyear{Hiddleston05}.

\xam\label{xam:bogus}
Assassin is in possession of a lethal poison, but has a last-minute
change of heart and refrains from putting it in Victim's coffee.
Bodyguard puts antidote in the coffee, which would have neutralized the
poison had there been any.  Victim drinks the coffee and survives.  
Is Bodyguard's putting in the antidote a cause of Victim surviving?
Most people would say no, but
according to the preliminary HP definition, it is.  For in the contingency
where Assassin puts in the poison, Victim survives iff Bodyguard puts
in the antidote.  
\exam

Example~\ref{xam:bogus} illustrates an even deeper problem with
Definition~\ref{actcaus}.  The structural equations for
Example~\ref{xam:bogus} are \emph{isomorphic} to those in the forest-fire
example, provided that we interpret the variables appropriately.
Specifically, take the endogenous variables in Example~\ref{xam:bogus} to be
$A$ (for ``assassin does not put in poison''), $B$ (for ``bodyguard puts
in antidote''), and $\VS$ (for ``victim survives'').
Then $A$, $B$, and $\VS$ satisfy exactly the same
equations as $L$, $\ML$, and $\FF$, respectively.  In the context where
there is lightning and the arsonists drops a lit match, both the 
the lightning and the match are causes of the forest fire, which seems
reasonable.  But here it does not seem reasonable that Bodyguard's
putting in the antidote is a cause.  Nevertheless, any
definition that just depends on the structural equations is bound to give
the same answers in these two examples.  (An example illustrating
the same phenomenon is given by Hall \citeyear{Hall07}.)  This suggests
that there must  
be more to causality than just the structural equations.  And, indeed, 
the final HP definition of causality allows certain contingencies to be
labeled as ``unreasonable'' or ``too farfetched''; these contingencies
are then not considered in AC2(a) or AC2(b).  Unfortunately, it is not
always clear what makes a contingency unreasonable.  Moreover, this
approach will not work to deal with Example~\ref{xam4}.

In this example, we clearly want to consider as reasonable the
contingency where no doctor is assigned to Billy and Billy is not
treated (and thus is sick on Tuesday).
We should also consider as reasonable the
contingency where doctor $1$ is assigned to Billy and treats him 
(otherwise we cannot say that doctor 1 is the cause of Billy
being sick if he is assigned to Billy and does not treat him). 
What about the contingency where doctor $i > 1$ is assigned to treat Billy and
does so?  It seems just as reasonable as the one where doctor 1 is
assigned to treat Billy and does so.  Indeed, if we do not call it
reasonable, then we will not be able to say that doctor $i$ is a cause of
Billy's sickness in the context where doctor $i$ assigned to treat Billy
and does not.  On the other hand, if we call it reasonable, then if
doctor 1 is assigned to treat Billy and does not, then doctor $i > 1$
not treating Billy will
also be a cause of Billy's sickness.   To deal with this, 
what is reasonable will have to depend on the context; in the
context where doctor 1 is assigned to treat Billy, it should not be
considered reasonable that doctor $i>1$ is assigned to treat Billy.

%Definition~\ref{actcaus} was labeled a preliminary definition of
%causality.  The final definition of causality given by Halpern and
%Pearl \citeyear{HP01b} tried to capture  the intuition that not all
%contingencies  were reasonable.  But it did not explicitly view the
%unreasonableness as coming from the reluctance to change the value of a
%variable from a ``normal'' value to an ``atypical'' value.
%Rather, it just extended the definition of causal model so as to include
%a list of allowable contingencies.   

As suggested in the introduction, the solution involves assuming that
an agent has, in addition to a theory of causality
(as modeled by the structural equations), a theory of ``normality'' or
``typicality''.  This theory would include statements like ``typically,
people do not put poison in coffee'' and ``typically doctors do not
treat patients to whom they are not assigned''.
%Here I propose a different modification of causal models.
%Motivated by Kahnemann and Miller's observation, I assume that the world
%is described not just by structural equations, but by ``typicality''
%statements.  
There are many ways of giving semantics to such
\fullv{
typicality statements, including {\em preferential structures\/}
    \cite{KLM,Shoham87}, {\em $\epsilon$-semantics\/}
    \cite{Adams:75,Geffner92,Pearl90}, and {\em possibilistic structures\/}
    \cite{DuboisPrade:Defaults91}, and ranking functions 
\cite{Goldszmidt92,spohn:88}.  For definiteness, I use the last approach
here (although it would be possible to use any of the other approaches
as well).}
\shortv{typicality statements (e.g., \cite{Adams:75,KLM,spohn:88}.  For definiteness, I use \emph{ranking
functions} \cite{spohn:88} here.}

Take a \emph{world} to be a complete description of the values
of all the random variables.  I assume that each world has associated
with it a \emph{rank}, which is just a natural number or $\infty$.
Intuitively, the higher the rank, the less likely the world.
A world with a rank of 0 is reasonably likely, one with a rank of 1 is
somewhat likely, one with a rank of 2 is quite unlikely, and so on.
Given a ranking on worlds, the statement ``if $p$ then typically $q$''
is true if in all the worlds of least rank where $p$ is true, $q$ is
also true.  Thus, in one model where people do not typically put either
poison or antidote in coffee, the worlds where neither poison nor
antidote is put in the coffee have rank 0, worlds where either
poison or antidote is put in the coffee have rank 1, and worlds
where both poison and antidote are put in the coffee have rank 2.

Take an {\em extended causal model\/} to
be a tuple $M = (\S,\F,\kappa)$, where $(\S,\F)$ is a causal model, and
$\kappa$ is a \emph{ranking function} that associates with each world a rank.
In an acyclic extended causal model, a context $\vec{u}$ determines a
world denoted $s_{\vec{u}}$.  
%In an extended model $M$, 
%a contingency $\vec{X} = \vec{x}'$; $\vec{W} =
%\vec{w}$ is \emph{acceptable relative to context $\vec{u}$} if there is
%a world $w'$ of lower rank than $w_{\vec{u}}$ 
%contingency (that is, $Y \in \vec{X} \union \vec{W}$ and $y$ is the value
%to which $Y$ is set), $Y=y$ is true in either
%$w_{\vec{u}}$ or $w'$.  That is, a contingency is acceptable relative to
%$\vec{u}$ if there is a world $w'$ of lower rank than $w_{\vec{u}$ such
%that if value of variable $Y$ in the contingency is different from its
%value in $w_{\vec{u}}$, then the change is to its value in a more normal
%world, namely $w'$.  
%We can now define what it means for 
$\vec{X}=\vec{x}$ is a \emph{cause of $\phi$
in an extended model $M$ and context $\vec{u}$} if $\vec{X}=\vec{x}$ is
a cause of 
$\phi$ according to Definition~\ref{actcaus}, except that in AC2(a), 
there must be a world $s$ such that $\kappa(s) \le \kappa(s_{\vec{u}})$
and $\vec{X} = \vec{x}' \land \vec{W} = \vec{w}$ is true at $s$.
This can be viewed as a formalization of Kahnemann and
Miller's observation that we tend to alter the  exceptional than the 
routine aspects of a world; we consider only alterations that
hold in a world that is no  more exceptional than the actual world.%
\footnote{I originally considered requiring that $\kappa(s) <
\kappa(s_{\vec{u}})$, so that you move to a strictly more normal world,
but this seems too strong a requirement.  
For example, suppose that $A$ wins an election over 
$B$ by a vote of 6--5.  We would like to say that each voter for $A$ is
a cause of $A$'s winning.  But if we view all voting patterns as equally
normal, then no voter is a cause of $A$'s winning, because no
contingency is more normal than any other.}
(The idea of extending causal models
with a ranking function already appears in \cite{HPearl01a}, but it was
not used to capture statements about typicality as suggested here.
Rather, it was used to talk about $\vec{X} = \vec{x}$ being a cause of
$\phi$ \emph{at rank $k$}, where $k$ is the lowest rank of the world
that shows that $\vec{X} = \vec{x}$ is a cause. 
The idea was dropped in the journal version of the paper.)

This definition deals well with all the problematic examples in the
literature.  Consider Example~\ref{xam:bogus}.  Using
the ranking described above, Bodyguard is not a cause of Victim's
survival because the world that would need to be considered in AC2(a),
where Assassin poison the coffee, is less normal than the actual world,
where he does not.  
It also deals well with Example~\ref{xam4}.
Suppose that in fact the hospital has 100
doctors and there are variables $A_1, \ldots, A_{100}$ and $\MT_1,
\ldots, \MT_{100}$ in the causal 
model, where $A_i = 1$ if doctor $i$ is assigned to treat Billy
\shortv{and $\MT_i = 1$ if he does, for $i = 1, \ldots, 100$.}
\fullv{and $A_i = 0$ if he is not, and
$\MT_i = 1$ if doctor $i$ actually treats Billy on Monday, and
$\MT_i = 0$ if he does not.}  
Doctor 1 is assigned to treat Billy; the
others are not.  However, in fact, no doctor treats Billy.  
Further assume that typically, doctors do not treat patients (that is, a
random doctor does not typically treat a random patient), and if doctor
$i$ is assigned to Billy, then typically doctor $i$ treats Billy.
%and if $i$ is assigned to Billy and $i$ does not treat Billy, then typically
%$i+1$ treats Billy (think of doctor $i+1$ as doctor $i$'s backup). I
%identify doctor 101 with doctor 1).
We can capture this in an extended causal model where the world
where no doctor is assigned to Billy and no doctor treats him has rank 0;
the 100 worlds where exactly one doctor is assigned to Billy, and that
doctor treats him, have rank
1; the 100 worlds where exactly one doctor is assigned to Billy and
%but doctor $i+1$ treats him 
no one treats him have rank 2;
%the world where $i$ is assigned to
%Billy but no doctor treats him rank 3, 
and the $100\times 99$ worlds where exactly one doctor is
assigned to Billy but some doctor 
%or $i+1$ 
treats him have
rank 3.  (The ranking given to other
worlds is irrelevant.)  In this extended model, in the context where
doctor $i$ 
is assigned to Billy but no one treats him, $i$ is the cause of Billy's
sickness  
%as is doctor $i+1$ 
(the world where $i$ treats Billy 
%or where $i+1$ treats Billy both have 
has 
lower rank than the world where $i$ is assigned to Billy but no one
treats him), but no other doctor is a cause of Billy's sickness.
Moreover, in the context where $i$ is assigned to Billy and treats him,
then $i$ is the cause of Billy's recovery (for AC2(a), consider the world
where no doctor is assigned to Billy and none treat him).  
%Similarly, if
%$i$ is assigned to Billy and does not treat him, but $i+1$ does, then
%$i+1$ is the cause of Billy's recovery.  None of 
%This cannot be captured in
%HP extended models, where some contingencies are declared allowable and
%some are not.  It is critical that what is allowable depends on the
%actual context.

\fullv{
I consider one more example here, due to Hitchcock
\citeyear{Hitchcock07}, that illustrates the interplay between normality
and causality.   

\xam\label{xam:Hitchcock}  Assistant Bodyguard puts a harmless antidote
in Victim's coffee.  Buddy then poisons the coffee, using a type of
poison that is normally lethal, but is countered by the antidote.
Buddy would not have poisoned the coffee if Assistant had not
administered the antidote first.  (Buddy and Assistant do not really
want to harm Victim.  They just want to help Assistant get a promotion
by making it look like he foiled an assassination attempt.)
Victim drinks the coffee and survives.  
\exam

Is Assistant's adding the antidote a cause of Victim's survival? 
Using the preliminary HP definition, it is; if Assistant does not add
the antidote, Victim survives.  However, using an extended causal
model with the normality assumptions implied by the story, it is not.
Specifically, suppose we assume that 
%normally Buddy does not poison the
%coffee and Assistant does not add the antidote, but 
if Assistant does
not add the antidote, then Buddy does not normally add poison. (Buddy,
after all, is normally a law-abiding citizen.)  In the corresponding
extended causal model, the world where Buddy poisons the coffee and
Assistant does not add the Antidote has a higher rank (i.e., is less
normal than) the world where  Buddy poisons the coffee and Assistant
adds the antidote.   This is all we need to know about the ranking
function to conclude that adding the antidote is not a cause.
By way of contrast, if Buddy were a more typical assassin, with
reasonable normality assumptions, the
world where he puts in the poison and Assistant puts in the antidote
would be less normal than then one Buddy puts in the poison and Assistant
does not put in the antidote, so Assistant would be a cause of Victim
being a alive.  

Interestingly, Hitchcock captures this story using structural equations
that also make Assistant putting in the antidote a \emph{cause} of
Buddy putting in the poison.  This is the device used to distinguish
this situation from one where Buddy is actually means Victim to die (in
which case Buddy would presumably have put in the poison even if Assistant
had not added the antidote).  However, it is not clear that people would
agree that Assistant putting in the antidote really \emph{caused} Buddy
to add the poison; rather, it set up a circumstance where Buddy was
willing to put it in.  I would argue that this is better captured by
using the normality statement ``If Assistant does not put in the
antidote, then Buddy does not normally add poison.''  As this example
shows, there is a nontrivial interplay between statements of causality
and statements of normality.
}
%\end{fullv}

I leave it to the reader to check that
reasonable assumptions about typicality can also be used to deal with
the other problematic examples for the HP definition that have been
pointed out in the literature, such as 
Larry the Loanshark \cite[Example 5.2]{HP01b} and 
Hall's \citeyear{Hall07}  watching police example.  (The family sleeps
peacefully through the night.  Are the watching police a cause?  After
all, if there had been thieves, the police would have nabbed them,
and without the police, the family's peace would have been disturbed.)

This is not the first attempt to modify structural equations to deal
with defaults; Hitchcock \citeyear{Hitchcock07} and Hall
\citeyear{Hall07} also consider this issue.
Neither adds any extra machinery such as ranking functions,
but both assume that there is an implicitly understood notion of
normality.  Roughly speaking, Hitchcock \citeyear{Hitchcock07} can be
understood as giving constraints on models that 
guarantee that the answer obtained using the preliminary HP definition
agrees with the answer obtained using the definition in extended causal
models.  I do not compare my suggestion to that of Hall
\citeyear{Hall07}, since, as Hitchcock \citeyear{Hitchcock07a} points
out, there are a number of serious problems with Hall's approach.
It is worth noting that both Hall and Hitchcock assume that a
variable has a ``normal'' or ``default'' setting; any other setting is
abnormal.  However, it is easy to construct examples where what counts
as normal depends on the context.  For example, it is normal for doctor
$i$ to treat Billy if $i$ is assigned to Billy; otherwise it is not. 

\commentout{
\section{Responsibility and blame}\label{sec:responsibility}
The HP definition of causality treats causality as an all-or-nothing
concept (as do all the other definitions of causality in the literature
that I am aware of).  While we can talk about the probability that $A$
is a cause of $B$ (by putting a probability on contexts), we cannot talk
about degree of causality.  This means that we cannot make some
distinctions that seem intuitively significant.  For example, as we
observed earlier, there seems to be a difference in the degree of
responsibility of a voter for a victory in an 11--0 election and a 6--5
election.   As Hana Chockler and I showed, one of the advantages of the
HP definition is that it provides a straightforward way of defining
refinements of the notion of causality that let us capture important
intuitions regarding degree of responsibility and blame.  The discussion
in this section is taken almost verbatim from \cite{ChocklerH03}.

\subsection{Responsibility}

The idea behind the definition of degree of responsibility is
straightforward:  If $A$ is not a cause $B$ then the degree 
of responsibility of $A$ for $B$ is 0.  If $A$ is a cause of $B$, then
the degree of responsibility of $A$
for $B$ is $1/(N+1)$, where $N$ is the minimal number of changes that
have to be made to obtain a contingency where $B$ counterfactually 
depends on $A$.  In the case of the 11--0 vote, the degree
of responsibility of any voter for the victory is $1/6$, since 5 changes
have to be made before a vote is critical.  If the vote were
1001--0, the degree of responsibility of any voter would be $1/501$.  On
the other hand, if the vote is 6--5, then the degree of responsibility
of each voter for is 1; each voter is critical.  

Thus, the degree of responsibility of $A$ for $B$ is a number between 0
and 1.  It is a refinement of the notion of causality.  It is 0 if and
only if $A$ is not a cause of $B$.  If $A$ is a cause of $B$, then the
degree of responsibility is positive.
Despite being a number between 0 and 1, as the voting examples make
clear, degree of responsibility does not act like probability at all.  

Here is the formal definition of degree of responsibility, from
\cite{ChocklerH03}.  

\dfn\label{def-resp}
The {\em degree of responsibility
of $\vec{X}=\vec{x}$ for $\phi$ in 
$(M,\vec{u})$\/}, denoted $\dr((M,\vec{u}), (\vec{X}=\vec{x}), \phi)$, is
$0$ if $\vec{X}=\vec{x}$ is 
not a cause of $\phi$ in $(M,\vec{u})$; it is $1/(k+1)$ if
$\vec{X}=\vec{x}$ is  a cause of $\phi$ in $(M,\vec{u})$ 
and there exists a partition $(\vec{Z},\vec{W})$ and an allowable setting%
\footnote{Here I am implicitly assuming that $M$ is an extended causal
model, so that it includes a set of allowable settings.}
$(\vec{x}',\vec{w})$ for which AC2 holds
such that (a) $k$ variables in $\vec{W}$ have different values in $\vec{w}$
than they do in the context $\vec{u}$ and (b) there is no partition 
$(\vec{Z}',\vec{W}')$ and setting $(\vec{x}'',\vec{w}')$ satisfying AC2
such that only $k' < k$ variables have different values in $\vec{w}'$
than they do in $\vec{u}$.
\edfn

It should be clear that the degree of responsibility for the voting
examples is indeed 1/6 in the case of an 11--0 victory and 1 in the case
of a 6--5 victory.   It is easy to see that in the context $(1,1,1)$ in
the forest fire example, where the lightning strikes, the arsonist drops
the match, and either one suffices for a fire, it is easy to see that
the lightning and the arsonist each have degree of responsibility $1/2$
for the fire.  On the other hand, in the context $(1,1,2)$, where both
are needed for the fire, then the lightning and the arsonist each have
degree of responsibility 1.  Finally, in the Suzy-Billy example, since
Billy is not a cause, Billy has degree of responsibility 0.   Suzy's
degree of responsibility depends on which settings are allowable in the
extended causal model.  
If we take $\vec{W}$ to consist of $\{\BT, \BH\}$,
and keep both variables at their actual setting in the context, 
so that $\BT=1$ and 
$\BH=0$, then Suzy's throw becomes
critical; if she throws, the bottle shatters, and if she does not throw,
the bottle does not shatter (since $\BH=0$).  On the other hand, if 
the setting $(\ST=0,\BT=1,\BH=0)$ is not allowable in the extended
causal model (on the grounds that it requires Billy's throw to miss),
but the arguably more reasonable settings $(\ST=0, \BT=0,\BH=0)$ and
$(\ST=1,\BT=1,\BH=0)$ where Billy does not throw are allowable, then
Suzy's degree of responsibility is $1/2$, since we
must consider the contingency where Billy does not throw.
Thus, using an appropriate extended causal model allows us to capture what
seems like a significant intuition here.

While the notion of degree of responsibility seems important, and
defining it in terms of the number of changes needed to make a
variable critical captures some of the intuitions we have, this is
admittedly a naive definition.  In some contexts it seems to come closer
to our intuitions to have the degree of responsibility of $\vec{X} =
\vec{x}$ decrease exponentially with the number of changes needed to
make $\vec{X} = \vec{x}$ critical, so that the degree of responsibility
would be, say, $1/2^k$ rather than $1/(k+1)$ if $k$ changes are needed.%
\footnote{I thank Denis Hilton for this suggestion.}
It may also be appropriate to assign weights to variables.  To
understand the intuition for this, consider a variant of the voting
example, where there are two voters, and each one controls a block of
votes: voter 1 
controls 8 votes and voter 2 controls 3 votes.  Moreover, each voter
must cast all his votes for one candidate; votes cannot be split. (This
is like voting in the U.S. Electoral College.  If we
consider a ``voter'' as representing a state in the Elctoral College,
for all states besides Nebraska and Maine, vote splitting is not
allowed; the winner of the popular vote in the state gets all the state's
electoral votes.)  If both voters vote for Mr. B, then only voter 1 is
the cause of Mr. B's victory, and thus has degree of responsibility 1.
On the other hand, if vote splitting is allowed (so that voter 1 can be
represented by a random variable that takes on values 0, \ldots, 8,
while voter 2 can be represented by a random variable that takes on
values 0, \ldots, 3), then both voter 1 and voter 2 are causes of
Mr. B's victory; however, voter 1 has degree of responsibility 1 (his
vote is clearly critical), while  voter 2 has degree of responsibility
$1/2$ (since voter 2 becomes critical if, for example,  voter 1 splits
his vote 4--4).  Note that, if vote splitting is allowed, voter 2 would
continue to have degree of responsibility $1/2$ even if he controlled
only one vote and voter 1 controlled 1,000 votes.  

Some might think 
that voter 2 should have a lower degree of responsibility, which takes
into account the fact that he controls fewer votes.  We could capture
this by assigning different weights to voters (or, more precisely, to
the variables representing), and having the definition of degree of
responsibility use this weight (rather than implicitly weighting all
voters equally).  That is, the degree of responsibility of
$\vec{X}=\vec{x}$ for $\phi$ would be 1 over 1 + the sum of the weights
of the variables that need to be changed to make $X = x$ critical.
While this would give voter 2 lower responsibility, it 
requires a modeler to assign a weight to variables.
Moreover, it is not so clear that weighting the variables captures all
our intuitions here.  
For example, suppose that, as before, voter 1 controls 8 votes, voter 2
controls 3 votes, but now there is a third voter that controls 10 votes.
If voters 1 and 2 vote for Mr.~B and voter 3 votes for Mr.~G, many would
agree that it is reasonable to assign both voters 1 and 2 degree of
responsibility 1 for the outcome.  Despite these concerns, this approach
has been found useful in verifying \emph{model checking} (an approach to 
verifying the correctness of computer programs) \cite{CGY08,CHK}.

An alternative that would not require any additional
information from the modeler would be to have the degree  of
responsibility of $\vec{X}=\vec{x}$ for $\phi$ depend on how many different
changes make $\vec{X}=\vec{x}$ critical.  For example, if voter 1
controls 8 votes and voter 2 controls 3, under this approach, voter 1
still has degree of responsibility 1, because all changes to voter 2's
votes make voter 1 
critical.  On the other hand, only three of the eight possible changes
to voter 1's vote (making the split 3--5, 4--4, or 5--3) make voter 2
critical.  Thus, voter 2's degree of responsibility would be $3/8$.  I
have not explored the implications of this modification of 
the definition.   While in the rest of the paper I use the
definition of degree of responsibility in Definition~\ref{def-resp}, a
variant may well be more appropriate in some applications.

\subsection{Blame}\label{sec:blame}
The definitions of both causality and
responsibility  are relative to an extended causal model and a context.
Thus, they implicitly assume that both are
given; there is no uncertainty.  Once we have a probability on contexts 
and causal models, we can talk about the probability of causality and
the expected degree of responsibility.  The latter notion is called 
\emph{(degree of) blame} by Chockler and Halpern.

Obviously, if we add probability to the picture, we must address the
question of where the probability is coming from.  In some cases it
might be objective; that is, it might come from a source with an
agreed-upon probability, like a coin toss.  Or we may have statistical
information that determines the probability.  Alternatively, the
probability may represent the agent's subjective beliefs.  As we shall
see, this latter interpretation is particularly important in some legal
notions.  For example, 
\emph{liability} depends on an agent's intentions and beliefs.  An agent
is not liable for a bad outcome if the agent did not believe that the
bad outcome would result from his actions (even though the
agent may be the cause of the outcome and, indeed, have degree of
responsibility 1).

To take a simple of how probability can be used,
suppose that an agent is unsure as to whether the context in
the forest fire examples is (1,1,0), (1,0,1), (1,1,1), or (1,1,2), and assigns
each of these contexts has probability $1/4$.   (For now let us not
worry about where the probability is coming from.)  With these
probabilities, the 
probability that 
lightning is a cause of fire is $3/4$ (since the lightning is a cause in
every context other than (1,1,0), where the fire was caused by something
other than the lightning or dropped match.  Note that 
the degree of responsibility of the lightning is 0 in context
(1,1,0), 1 in the contexts (1,0,1) and (1,1,2), and $1/2$ in the context
(1,1,1), where either the lightning or the dropped match suffices to
cause the forest fire.  Thus, the degree of blame (i.e., expected degree
of responsibility) is $5/8$ ($= - 0 \times 1/4 + 1 \times 1/4 + 1/2
\times 1/4 + 1 \times 1/4$).

As I said earlier,
there are two significant sources of uncertainty for an
agent who is contemplating performing an action:
\begin{itemize}
\item 
what the true situation is (that is, what value the exogenous variables
have); for example, a doctor may be 
uncertain about whether a patient has high blood pressure;
\item how the world works; for example, a doctor may be uncertain about
the side effects of a given medication.
\end{itemize}

In the HP framework, the ``true situation'' is determined by the context;
``how the world works'' is determined by the equations.
All the uncertainty about the equations can be encoded into the
context.  
%This is the case in the causal model for the
%forest fire example.  The question of whether a
%fire needs both a dropped match and a lightning strike, or whether one
%of these is enough, can be eoncded by using two different causal models
%(which is what was actually done in \cite{HP01b}).  Here it is encoded
%into the context.  If the third component of the context is 1, then the
%forest fire will start if either the match is dropped or lightning
%strikes, while if the third component is 2, then both are required.
For example, in modeling the forest fire, I used the context $U$ to
describe whether both the match and the lightning are needed for the
fire, or whether one of them suffices.
But at times it
is convenient to simply use two different causal models.  For example,
we can imagine a causal model for cancer where smoking causes cancer,
and another causal model where some genetic problem results in a greater
likelihood of cancer and a greater propensity to smoke.  While we now
believe that the first causal model more accurately depicts reality, at
one point there was some doubt.

In any case, motivated by this discussion, I take an agent's
uncertainty to be modeled by a pair  $(\K,\Pr)$, where $\K$ 
is a set of \emph{situations}, pairs of the form $(M,\vec{u})$,
where $M$ is an extended
causal model and $\vec{u}$ is a context,
and $\Pr$ is a probability distribution over $\K$.
In the forest fire example, in all the situations, the causal model was
the same, but in general this need not be the case.  
Intuitively, $\K$ describes the situations that the agent considers
possible before $\vec{X}$ is set to $\vec{x}$.
(Note that the situation $(M_{\vec{X} \gets \vec{x}},\vec{u})$ for $(M,
\vec{u}) \in \K$  are those 
that the agent considers possible after $X$ is set to $x$.)  
The degree  of blame that setting $\vec{X}$ to $\vec{x}$ has for $\phi$ is 
then the expected degree
of responsibility of $\vec{X}=\vec{x}$ for $\phi$ in 
$(M_{\vec{X} \gets x},\vec{u})$,   
taken over the situations $(M,\vec{u}) \in \K$.

\dfn
The {\em degree of 
blame of setting $\vec{X}$ to $\vec{x}$ for $\phi$ relative to epistemic state
$(\K,\Pr)$\/}, denoted $\db(\K,\Pr,\vec{X} \gets \vec{x}, \phi)$, is
$$\sum_{(M,\vec{u}) \in \K}
\dr((M_{\vec{X} \gets \vec{x}}, \vec{u}), \vec{X} = \vec{x}, \phi)
\Pr((M,\vec{u})).$$ 
\end{definition} 
In this definition, it is perhaps best to think of $\vec{X}=\vec{x}$ as indicating
that some action has been performed.  Thus, if we are trying to decide
to what extent an agent who performs a particular action is to blame for
an outcome, we can take $X$ to be a random variable that indicates
whether the action is performed (so that $X=1$ if the action is
performed and $X=0$ otherwise) and $\phi$ to be the outcome of
interest.  To determine the degree of blame attached to $X \gets x$, we first
consider what situations the agent considers possible \emph{before} the
action is performed, and how likely each one of them is (according to
the agent).  This is given by $(\K,\Pr)$.  We then consider the agent's
degree of responsibility in each model that arises if the action is
actually performed.  If $(M,u)$ is one of the situations the agent
considers possible before performing the action, after performing the
action, the situation is described by $M_{X \gets 1}$; we thus consider
the degree of responsibility of $X=1$ for the outcome $\phi$ in the
causal model $M_{X \gets 1}$.    It may also make sense to put a
probability on the situations that arise \emph{after} the action is
performed.  I return to this issue below, after considering a few examples.

\begin{example} Suppose that we are trying to compute the degree
of blame of Suzy's throwing the rock for the bottle shattering. 
Suppose that the only causal model that Suzy considers possible is
essentially like the second model in Example~\ref{xam2} (with $\SH$ and
$\BH$), with some minor modifications: 
$\BT$ can now take on three values, say 0, 1, 2; as before, if
$\BT = 0$ then Billy doesn't throw, if $\BT = 1$, then Billy does throw,
and if $\BT = 2$, then Billy throws extra hard.  Assume that the causal
model is such that if $\BT = 1$, then Suzy's rock will hit the bottle
first, but if $\BT =2$, they will hit simultaneously.  Thus, $\SH = 1$
if $\ST = 1$, and $\BH = 1$ if $\BT = 1$ and $\SH = 0$ or if $\BT = 2$.
Call this structural model $M$.

At time 0, Suzy considers the following four situations equally likely:
\begin{itemize}
%\item $(M_1, \vec{u}_1)$, where the bottle was already shattered before
%Suzy's throw; 
%\item $(M_2,\vec{u}_2)$, where the bottle was whole before Suzy's throw, 
%and Suzy and Billy both hit the bottle simultaneously 
%(as described in the model in Figure~\ref{fig0});
%\item $(M_3,\vec{u}_3)$, where the bottle was whole before Suzy's throw,
%and Suzy's throw hit before Billy's throw (as described in the model in
%Figure~\ref{fig1});
%\item $(M_4,\vec{u}_4)$, where the bottle was whole before
%Suzy's throw, and Billy did not throw. 
\item $(M, \vec{u}_1)$, where $\vec{u}_1$ is such that Billy already
threw at time 0 (and hence the bottle is shattered);
\item $(M,\vec{u}_2)$, where the bottle was whole before
Suzy's throw, 
and Billy throws extra hard, so 
Billy's throw and Suzy's throw hit the bottle
simultaneously (this essentially gives the first model in
Example~\ref{xam2});
\item $(M,\vec{u}_3)$, where the bottle was whole before Suzy's throw,
and Suzy's throw hit before Billy's throw (this essentially gives the
second model in Example~\ref{xam2}); and
\item $(M,\vec{u}_4)$, where the bottle was whole before
Suzy's throw, and Billy did not throw. 
\end{itemize}
The
bottle is already shattered in $(M,\vec{u}_1)$ before Suzy's
action,
so Suzy's throw is not a cause of the bottle shattering, and her degree
of responsibility for the shattered bottle is 0.
As discussed earlier,
the degree of 
responsibility of Suzy's throw for the bottle shattering is $1/2$ in
$(M,\vec{u}_2)$ and 1 in both $(M,\vec{u}_3)$ and $(M,\vec{u}_4)$.
Thus, the degree of blame is $\frac{1}{4}\cdot \frac{1}{2} +
\frac{1}{4}\cdot 1 + \frac{1}{4}\cdot 1 = \frac{5}{8}$.  
If we further require that the contingencies in AC2(b) involve only
allowable settings, and assume that the setting $(\ST=0, \BT=1, \BH=0)$ is not
allowable, then the degree of responsibility of Suzy's throw in
$(M,\vec{u}_3)$ is $1/2$; in this case, the degree of blame is
$\frac{1}{4}\cdot \frac{1}{2} + \frac{1}{4}\cdot \frac{1}{2} +
\frac{1}{4}\cdot 1 = \frac{1}{2}$.  

Note that if we consider Suzy's probability on situations after the rock
is thrown and Suzy observes what happens, then she knows the outcome
with probability 1.  Thus, her degree of blame is exactly her degree of
responsibility,  If Suzy is considering whether to throw the rock and
wants to consider how much she will be to blame if the bottle shatters,
it seems more appropriate to consider her prior probability on
situations.  If we want to consider her degree of blame given what has
happened, then it makes more sense to consider the posterior probability.
\end{example}

\begin{example}\label{xam:firingsquad}
Consider a firing squad with ten  excellent marksmen.  Suppose that
marksman 1 knows that exactly one 
marksman has a live bullet in his rifle, 
and that all the marksmen will shoot.
 Thus, he considers 10
situations possible, depending on who has the bullet.  Let $p_i$ be some
marksman 1's prior probability that marksman $i$ has the live
bullet.  In situation $i$, marksman $i$ is the cause of death and has
degree of responsibility 1; in all other situations, marksman $i$ is not
the cause of death and has degree of responsibility 0.  Thus, 
the probability that marksman $i$ is the cause of death is
$p_i$, and marksman $i$'s degree of blame is also $p_i$. 
Note that if marksman 1 mistakenly believes that he has
the bullet (and thus takes $p_1=1$) when in fact it is marksman 2, then
it is possible for the degree of blame of marksman 1 (according to marksman
1) to be 1, while in fact the degree of responsibility of
marksman 1 is 0.  This shows that degree of blame is a subjective
notion, depending on an agent's subjective probability.

Note that if the marksman never actually discovers which bullet was
live, then his prior probability is the same as the posterior
probability.  It does not matter which we use to compute the degree of
blame.  If he discovers which bullet is live, then his degree of blame
will be equal to the degree of responsibility.  Finally, if is given 
only partial information about which bullet is live, then it is
straightforward to compute an appropriate degree of blame based on his
posterior probability.
\end{example}

%Example~\label{xam:firingsquad} suggests that both degree of blame and
%degree of responsibility may both be relevant in a legal setting.  An
%agent's beliefs are relevant when judging intent.  The actual situation
%is important for judging actual causality.

In legal cases, it may not always be appropriate to consider an agent's
actual epistemic state for the probability.  It may be more appropriate
to consider what the epistemic state should have been.  The definition
of blame does not change; the definition is agnostic as to what
epistemic state should be considered.  Considering the actual epistemic
state is relevant when considering intent; what the epistemic state
should have been may be more appropriate in assessing liability.
Consider, for example, a patient who dies as a result of being treated
by a doctor with a particular drug.
Assume that the patient died due to the drug's adverse side effects 
on people with high blood pressure and, for simplicity, that this was
the only cause of death. Suppose that the doctor was not 
aware of the drug's adverse side
effects.  (Formally, this means that he does not consider possible a
situation with a causal model where taking the drug causes death.)
Then, relative to the doctor's actual epistemic state, 
the doctor's degree of blame will be 0.
However, a lawyer might argue in court that the doctor should  have
known that treatment had adverse side effects for patients with high
blood pressure (because this is well documented in the literature) 
and thus should have checked the patient's blood pressure.
If the doctor had performed this test, he would of course have known
that the patient had high blood pressure.  With respect to the resulting
epistemic state, the doctor's degree of blame for the death is quite
high.  Of course, the lawyer's job is to convince the court that the
latter epistemic state is the appropriate one to consider when assigning
degree of blame.  

As an orthogonal issue, we can also consider the posterior probability
here.  The doctor's epistemic state after the patient's death
is likely to be quite different from her epistemic state before the
patient's death.  She may still consider it possible that the patient
died for reasons other than the treatment, but will consider causal
structures where the treatment was a cause of death more likely.  Thus,
the doctor will likely have higher degree of blame relative to her 
epistemic state after the treatment. 

Interestingly, all three epistemic states---the epistemic state that an
agent actually has before performing an action, the epistemic state that
the agent should have had before performing the action, and the
epistemic state after performing the action---have been considered
relevant to determining responsibility according to different legal theories
\cite[p.~482]{HH85}.   I do not think that there is a single ``right''
definition here.  The appropriate epistemic state depends on the
application.  What is important is that the HP framework lets us explain
the differences clearly.  
}
   
\section{The NESS approach}\label{sec:NESS}

%As I said, there has been extensive research on causality in both
%philosophy and law.  Pearl and I \citeyear{HP01b} have already compared
%our approach to other work in the philosophy literature, so I focus here
%on work in the legal literature.  Perhaps the best worked-out approach is
%the NESS (Necessary Element of a Sufficient Set) test, originally
%described by Hart and Honor\'{e}, and worked out in much greater detail
%by Wright \citeyear{Wright85,wright:88,Wright01}.  Thus, I compare the
%HP approach to the NESS approach here.
In this section I provide a sufficient condition to guarantee that a
single conjunct is a cause.  Doing so has the added benefit of providing
a careful comparison of the NESS test and the HP approach.
Wright does not provide a mathematical formalization of the NESS test; 
what I give here is my understanding of it.  
%As I hope to show, there
%are some important subtleties in the definition that need
%clarification.  

$A$ is a cause of $B$ according to the NESS test if
there exists a set $\Suff = \{A_1, \ldots, A_k\}$ of events,
each of which 
actually occurred, where $A = A_1$, $\Suff$ is sufficient for 
for $B$, and $\Suff - \{A_1\}$ is not sufficient for $B$.  Thus, $A$ is an
element of a sufficient condition for $B$, namely $\Suff$,
and is a necessary element of that set, because any subset of
$\{A_1, \ldots, A_k\}$ that does not include $A$ is not sufficient for
$B$.%
\footnote{The NESS test
is much in the spirit of Mackie's INUS test \cite{mackie:65}, according
to which $A$ is a cause of $B$ if $A$ is an insufficient 
but necessary part of a condition which is unnecessary but sufficient
for $B$.  However, a comparison of the two  approaches is beyond the
scope of this paper.}

The NESS test, as stated, seems intuitive and simple.
Moreover, it deals well with many examples.
%Consider the forest fire.  The lightning and the arsonist are clearly
%both causes; we can take the set $\Suff$ to be the singleton set consisting
%of either lightning or the arsonist dropping a match.  Similarly, in 
%Example~\ref{xam4}, if both Monday's doctor treating Billy and
%Tuesday's doctor not treating Billy are elements of $\Suff$, then each of
%them are causes. However, I believe that the NESS 
%approach has problems with the Suzy-Billy example.  These are best
%pointed out by considering a related example also considered by
%Wright \citeyear{Wright85}.  This example shows that, 
However,  
%\citeyear{Wright85}) shows,
%However, the NESS approach suffers from some serious problems.
%While its 
although the NESS test looks quite formal, it lacks a 
definition of what it means for a set $\Suff$ of events to be
\emph{sufficient} for $B$ to occur. As I now show, such a definition is
sorely needed.  

\xam\label{xam:poison} 
Consider Wright's example of Victoria's poisoning from the introduction.
First, suppose that Victoria drinks a cup of tea
poisoned by Paula, and then dies.  It seems clear that Paula
poisoning the tea caused Victoria's death.  
Let $\Suff$ consist of two events:
\fullv{\begin{itemize}
\item}
$A_1$, Paula poisoned the tea; and 
\fullv{\item}
$A_2$, Victoria drank the tea.
\fullv{\end{itemize}}
Given our understanding of the world, 
%(which in the HP definition is encoded by the structural equations) 
it seems reasonable to say that the
$A_1$ and $A_2$ are sufficient for Victoria's death, but removing
$A_1$ results in a set that is insufficient.

But now suppose that Sharon shoots Victoria just after she drinks the tea
(call this event $A_3$),
and she dies instantaneously from the shot (before the poison can take
effect).  In this case, we would want to say that 
%Sharon's shot is the
$A_3$ is the
cause of Victoria's death, not $A_2$.  Nevertheless, it would 
seem that the same argument that makes Paula's poisoning a cause without
Sharon's shot would still make Paula's poisoning a cause even without
Sharon's shot.  The set $\{A_1,A_2\}$ still seems sufficient for
Victoria's death, while $\{A_2\}$ is not.  

Wright \citeyear{Wright85} observes the poisoned tea would be a cause of
Victoria's death only if Victoria ``drank the tea and \emph{was alive when the
poison took effect}''.   Wright seems to be arguing that $\{A_1,A_2\}$ is in
fact \emph{not} sufficient for Victoria's death.  We need $A_3$: Victoria was
alive when the poison took effect.  While I agree that the fact that
Victoria was alive when the poison took place is critical for causality, I
do not see how it helps in the NESS test, under what seems to me the
most obvious definitions of ``sufficient''.  I would argue that
$\{A_1,A_2\}$ \emph{is} in fact just as sufficient for death as
$\{A_1, A_2, A_3\}$.  For suppose that $A_1$ and $A_2$ hold.  Either
Victoria was alive when the poison took effect, or she was not.  In the
either case, she dies.  In the former case, it is due to the poison; in
the latter case, it is not.  
\shortv{\exam}
%Perhaps the NESS test should require that
%the set $\Suff$ be sufficient for the poisoning to cause the death,
%rather than 
%just being sufficient for death.  But in that case, it is even more
%crucial that there be a definition of what it means for a set to be
%sufficient for death by poisoning.  

\fullv{
But it gets worse.  While I would argue that
$\{A_1, A_2\}$ is indeed just as sufficient
for death as $\{A_1,A_2,A_3\}$, it is not clear that $\{A_1,A_2\}$ is in
fact sufficient.  Suppose, for example, that some people are
naturally immune to the poison that Paula used, and do not die from it.
Victoria is not immune.  But 
then it seems that we need to add a condition $A_4$ saying that Victoria
is not immune from the poison to get a set sufficient to cause Victoria's
death.  And why should it stop there?  Suppose that the poison has an
antidote that, if administered within five minutes of the poison taking
effect, will prevent death.  Unfortunately, the antidote was not
administered to Victoria, but do we have to add this condition to $\Suff$ to get a
sufficient set for Victoria's death?  Where does it stop?
\exam
}
%\end{fullv}

\commentout{
Note that in the causal model where the only random variables are ``Paula 
poisoned the tea'', ``Sharon shot Victoria'', ``Victoria was alive when the
poison took effect'', and ``Victoria dies'' (where the random variable has
value 1 or 0, depending on whether the event happened), with the
obvious equations, the shot is indeed a cause of death in the HP
definition, while the poisoning is not.  The argument is almost
identical to the Suzy-Billy case.  Moreover, adding additional random
variables like ``Paula is naturally immune'' or ``the antidote was
administered'' does not make a difference.   However, in general, adding
more variables might make a difference.  (Recall that, in the Suzy-Billy
example, adding the variables $\BH$ and $\SH$ was important, as is
``Victoria was alive when the poison took effect'' in this case.)  That is
why the causal model must make explicit what variables are being considered.
\exam
}

\journal{
The issue of what counts as a sufficient cause is further complicated if
there is uncertainty about the causal structure.  

\xam\label{medication} Suppose that a doctor gives a patient a new drug
to deal with a heart ailment, and then the patient dies.  Is the new
drug the cause of death.  Even ignoring all the issues raised in
Example~\ref{xam:poison}, it  is clear that the answer depends on
whether giving the drug is a sufficient condition to cause death (given
all the other factors affecting the patient).  The NESS seems to
implicitly assume that this is known.  For example, Wright
\citeyear{Wright06} says that the putative cause ``must be part of the
complete instantiation of the antecedent of the relevant causal law''.
But the notion of a causal law is undefined, nor is it defined when a
causal law is ``relevant''.  By way of contrast, in causal models, the
causal laws are encoded by the structural equations.
\exam
}

\nokr{
The NESS definition is also unclear as to which events can go in
$\Suff$.  The problem is illustrated in the next example.
%This already arises in the
%analysis of Example~\ref{xam4}.  Can $\Suff$ include ``negative'' events
%like ``Tuesday's doctor did \emph{not} treat Billy''.  If so, can it
%also include ``Doctor $i$ did not treat Billy'' for $i = 1, \ldots, 99$,
%for the other 99 doctors who did not treat Billy?  The next example
%shows that the problem is quite pervasive.

\commentout{
The next example shows that whether something is a cause may depend on 
whether $\Suff$ is allowed to contain disjunctive events.
\xam\label{xam:disjunctive} Suppose that a sensor senses the sum of the
forces applied to an object.  It reports a problem if the total force is
too high or too low.  For simplicity, suppose that the first force,
described by the random variable $F_1$, is either 0 or 1, while the
second force, is either 0, 1, or 2.  The sensor $S$ reads 1 (``good'')
if the total force $F_1 + F_2$ is either 1 or 2.  In the actual context,
$F_1 =0$ and $F_2 = 1$, so $S=1$.  According the HP model, both $F_1 =
0$ and $F_2 = 1$ are causes of $S=1$.  (To see that $F_1=1$ is a cause,
consider the contingency where $F_2 = 2$.)  Are they both causes
according to the NESS test?  Certainly $F_2 = 1$ seems to be; we can
just take $\Suff = \{F_2 = 1\}$.  What about $F_1 = 0$?  Certainly $F_1 = 0$
is not by itself sufficient for $S=1$.  But if $F_2 = 1$ is in $\Suff$, then
$F_1 = 0$ is no longer necessary.  However, if we take $\Suff$ to consist of
the disjunctive event $F_2 = 1 \lor F_2 = 2$ (which actually occurred)
and $F_1 = 0$, then this set is necessary for $S=1$, while $F_2 =1 \lor
F_2 = 2$ is not.  Thus, if disjunctive events are allowed in $\Suff$, then
$F_1 = 0$ is a cause of $S=1$.
\exam
}

\xam\label{xam:discharge} Wright \citeyear{Wright01} considers an
example where defendant 1 discharged 15 units of effluent, while defendant 2
discharged 13 units.  Suppose that 14 units of effluent are sufficient
for injury.  It seems clear that defendant 1's discharge is a cause of
injury; if he hadn't discharged any effluent, then there would have been
no injury.  What about defendant 2's discharge?  In the HP approach,
whether it is a cause depends on the random variables considered and their
possible values.  Suppose that $D_i$ is a random variable representing
defendant $i$'s discharge, for $i = 1, 2$.  If $D_1$ can only take
values 0 or 15 (i.e., if defendant 1 discharges either nothing or all 15
units), then defendant 2's discharge is not a cause.  But if $D_1$ can
take, for example, every integer value between 0 and 15, then $D_2 = 13$
is a cause (under the contingency that $D_1 = 4$, for example).

Intuitively, the decision as to whether the causal model should include
4 as a possible value of $D_1$ or have 0 and 15 as the only possible
values of $D_1$ should depend on the options available to defendant 1.
If all he can do is to press a switch that determines whether or not
there is effluent (so that pressing the switch results in $D_1$ being
15, and not pressing it result in $D_1$ being 0) then it seems
reasonable to take 0 and 15 as the only values.  On the other hand, if
the defendant can control the amount of effluent, then taking the range
of values to include every number between 0 and 15 seems more
reasonable.  

Perhaps not surprisingly, this 
issue is relevant to the NESS test as well, for the same reason.  If the
only possible values of $D_1$ are 0 or 15, then there is no set $\Suff$
including $D_2 = 13$ that is sufficient for the injury such that $D_2 =
13$ is necessary.  On the other hand, if $D_1 = 4$ is a possible event,
then there is such a set.
\exam

The problem raised by Example~\ref{xam:discharge}, that of which events
can go 
into $\Suff$, is easy to deal with, by simply making the set of
variables that can go into $\Suff$ explicit.
Of course, as the example suggests, the choice of events will
have an impact on what counts as a cause, but that is arguably
appropriate.  
Recall that causal models deal with this issue by making
explicit the signature, that is, the set of variables and their possible
values.  This gives us a set of 
primitive events of the form $X=x$.  More complicated events can be
formed as Boolean combinations of primitive events,  but it may also be
reasonable to restrict $\Suff$ to consisting of only primitive events.

The problem raised by Example~\ref{xam:poison}, that of defining
sufficient cause, seems more serious. I 
believe that a formal definition will require some of the machinery of
causal models, including structural equations.}
\kr{I believe that a formal definition of sufficient cause requires 
the machinery of causal models.}  (This point echoes
criticisms of NESS and related approaches by Pearl
\citeyear[pp. 314--315]{pearl:2k}.)  
%I have not been able to find an
%unproblematic way to completely define sufficiency, 
%even using causal models.   
I now sketch an approach to defining sufficiency that 
delivers reasonable answers in many cases of interest and, indeed,
often agrees with the HP definition.%
\footnote{Interestingly, Baldwin and Neufeld \citeyear{BN03} claimed
that the NESS test could be formalized using causal models, but did not
actually show how, beyond describing some examples.  In a later paper
\cite{BN04}, they seem to 
retract the claim that the NESS test can be formalized using causal
models.}

Fix a causal model $M$.  Recall that a primitive
event has the form $X=x$; a set of primitive events is \emph{consistent}
if it does not contain both $X=x$ and $X=x'$ for some random variable $X$ and
$x \ne x'$.  If $\Suff = \{X_1 = x_1, \ldots, X_k = x_k\}$ is a consistent
set of primitive events, then $\Suff$ is
\emph{sufficient} for $\phi$  relative to causal model $M$ if 
$M \sat [\Suff]\phi$, where $[\Suff]\phi$ is an abbreviation for
$[X_1 \gets x_1; \ldots; X_k \gets x_k] \phi$.  Roughly speaking, the
idea is to formalize the NESS test by taking $X=x$ to be a cause of
$\phi$ if there is a a set $\Suff$ including $X=x$ that is sufficient for
$\phi$, while $\Suff - \{X=x\}$ is not.  Example~\ref{xam:poison}
already shows that this will not work.  If $\CP$ is
a random variable that takes on value 1 if Paula poisoned the tea and 0
otherwise, then it is not hard to show that in the obvious causal model,
$\CP=1$ is sufficient for $\PD=1$ (Victoria dies), even if Sharon shoots
Victoria.  To deal with this problem, we must strengthen the notion of
sufficiency to capture some of the intuitions behind AC2(b). 

Say that $\Suff$ is \emph{strongly sufficient for $\phi$ in
$(M,\vec{u})$} if  $\Suff \union \Suff'$ is sufficient for $\phi$ 
in $M$ for all 
sets $\Suff'$ consisting of primitive events $Z=z$ such that $(M,\vec{u})
\sat Z=z$.  Intuitively, $\Suff$ is strongly sufficient for $\phi$ 
in $(M,\vec{u})$ if $\Suff$ remains sufficient for $\phi$ even when
additional events, which happen to be true in $(M,\vec{u})$, are added to
it.  
\commentout{
It may seem strange that a set $\Suff$ that is sufficient for
$\phi$ does not continue to be sufficient for $\phi$ as more events are
added to it.  But consider the Suzy-Billy example.  Given the structural
equations in that example, Billy throwing is sufficient for the bottle
shattering.  In all context where Billy throws, the bottle shatters
(assuming that the context just determines who throws and who does not).
But if $\BH=0$ is added to $\{\BT=1\}$, then the resulting set is
\emph{not} sufficient for $\BS=1$; if Billy's rock does not hit despite
Billy throwing, then the bottle does not  necessarily shatter.  Thus,
despite the fact that there is no context where Billy throws and the
bottle does not shatter, Billy throwing is not strongly sufficient for
the bottle shattering in the context where both Billy and Suzy throw
(i.e., in a context where $\BH=0$).
We can use a similar argument to show that $\CP = 1$ is not strongly
sufficient for $\PD=1$ in the actual context, provided that the language
includes enough events.}
%\end{commentout}
As I now show, although $\CP=1$ is sufficient for $\PD=1$, it is not strongly
sufficient, provided that the language includes enough events.
%Recall the moral from the Suzy-Billy story.  In

As already shown by HP, in 
order to get the ``right'' answer for causality in the presence of
preemption (here, the shot preempts the poison), there must be a
variable in the language that takes on 
different values depending on which of the two potential causes is the
actual cause.  In this case, we need a variable that takes on different
values depending on whether Sharon shot. 
%this variable  corresponds to $\BH$ in the Suzy-Billy story.  
Suppose that it
would take Victoria $t$ units of time after the poison is administered to
die; let $\AC$ be the variable that has value 1 if Victoria dies $t$ units
of time after the poison is administered and is alive before that, and
has value 0 otherwise.  Note that
$\AC=0$ if Victoria is already dead before the poison takes
effect.  In particular, if Sharon shoots Victoria before the poison takes
effect, then $\AC=0$.  
%But $\AC$ is also 0 in a context where Victoria is 
%alive before the poison starts taking effect and is still alive after
%the poison should have taken effect (perhaps because Victoria was given an
%antidote).   
Then 
%(assuming that there is no context where Victoria is immune to
%the poison), 
although $\CP=1$ is sufficient for $\PD=1$, it is not
strongly sufficient for $\PD=1$ in the context $\vec{u}'$ where Sharon 
shoots, since $(M,\vec{u}) \sat \AC = 0$, and
$M \sat [CP\gets 1; \AC \gets 0] (\PD \ne 1)$.

The following definition is my attempt at formalizing the NESS
condition, using the ideas above.

\dfn $\vec{X} = \vec{x}$ is a {\em cause of $\phi$ in
$(M, \vec{u})$ according to the causal NESS test} if there exists a set
$\Suff$ of primitive events containing $\vec{X}=\vec{x}$ such that 
the following properties hold:
\begin{description}
\item[{\rm NT1.}] $(M,\vec{u}) \sat \Suff$; that is, $(M,\vec{u}) \sat 
Y=y$ for all primitive events $Y=y$ in $\Suff$.
\item[{\rm NT2.}] $\Suff$ is strongly sufficient for $\phi$ in
$(M,\vec{u})$.
\item[{\rm NT3.}] $\Suff - \{\vec{X}=\vec{x}\}$ is not 
strongly
sufficient for $\phi$
in $(M,\vec{u})$.
\item[{\rm NT4.}] $\vec{X} = \vec{x}$ is minimal; no subset of $\vec{X}$
satisfies conditions NT1--3.%
\footnote{This definition does not take into account defaults.  It can
be extended to take defaults into account by requiring that if
$\vec{u}'$ is the context showing that $\Suff - \{X=x\}$ is not strongly
sufficient for $\phi$ in NT2, then $\kappa(s_{\vec{u}'}) \le 
\kappa(s_{\vec{u}})$.  For ease of exposition, I ignore this issue here.}
%\qed
\end{description}
$\Suff$ is said to be a \emph{witness} for the fact that $\vec{X} =
\vec{x}$ is a cause of $\phi$ according to the causal NESS test.
\edfn
%\shortv{\end{definition}}

Unlike the HP definition, causes according to the causal NESS
test
%\shortv{As I show in the full paper, unlike the HP definition, causes
%according to the causal NESS test} 
always consist of single conjuncts.

\thm\label{pro:singlecause} If $\{X_1 = x_1, \ldots, X_k = x_k\}$ is a
cause of $\phi$ in $M$ 
according to the causal NESS test, then $k=1$.
\ethm

%joe
%Since, as Example~\ref{xam3b} shows, a cause according to the HP
%definition might not be a single conjunct, the HP definition and the
%NESS definition are incomparable.    
It is easy to check that in Example~\ref{xam3b}, both $C_1=1$ and
$C_2=1$ are causes of $\WIN=1$ according to the causal NESS test, while 
(because of NT4) $C_1=1\land C_2=1$ is not.  On the other hand,
Example~\ref{xam3b} shows that neither $C_1=1$ nor $C_2=1$ is a cause
according to the HP definition, while $C_1 \land C_2=1$ is.  Thus, the
two definitions are incomparable.

%Moreover, at times, the requirement
%in the NESS test that strong sufficiency hold in \emph{all} contexts is
%clearly too strong.  

Nevertheless, the HP definition and the causal NESS test agree in many
cases of interest (in particular, in 
all the examples in the HP paper). 
In light of Theorem~\ref{pro:singlecause}, this explains in part why,
in so many cases, causes are single conjuncts with the HP definition.
In the rest of this section
I give conditions under which
the NESS test and the HP definition agree.
Although they are complicated, they
apply in all the standard examples in the literature.

%\fullv{
I start with conditions that suffice to show that being a
cause with according to the causal NESS test implies being a cause
according to the HP definition.

\thm\label{pro:NESSsuff} Suppose that $X=x$ is a cause of $\phi$ in
$(M,\vec{u})$ according 
to the causal NESS test with witness $\Suff$, and there exists a
(possible empty) set $\vec{T}$ of variables not mentioned in $\phi$ or
$\Suff$ 
and a context $\vec{u}'$ such that the following properties hold:
\begin{description}
\item[SH1.] $\Suff - \{X=x\}$ is not a sufficient condition for $\phi$
in $(M,\vec{u}')$; that is, $(M,\vec{u}') \sat [\Suff - \{X=x\}] \neg \phi$.%
\nokr{\footnote{Since $\Suff$ is a witness to the fact that $X=x$ is a cause
of $\phi$ in $(M,\vec{u})$, $\Suff - 
\{X=x\}$ is not a strongly sufficient cause for $\phi$ with respect to
$(M,\vec{u})$.  SH1 requires something different: that $\Suff -
\{X=x\}$ not be a sufficient cause for $\phi$ in $(M,\vec{u}')$.}
}
\item[SH2.] Each variable in $\vec{T}$ is independent of all other
variables in contexts $\vec{u}$ and $\vec{u}'$; that is, for all
variables $T \in \vec{T}$, if $\vec{W}$ consists of all endogenous
variables other than $T$, then for all settings $t$ of $T$ and 
$\vec{w}$ of $\vec{W}$, we have 
$(M,\vec{u}) \sat T = t$ iff $(M,\vec{u}) \sat
[\vec{W} \gets \vec{w}] (T = t)$, and similarly for
context $\vec{u}'$.
\item[SH3.] $\phi$ is determined by $\vec{T}$ and $X$ in contexts
$\vec{u}$ and $\vec{u}'$; that is, for all $\vec{t}$, $\vec{T}'$
disjoint from $\vec{T}$ and $X$, $x'$, and $\vec{t}'$,
we have  $(M,\vec{u}') \sat [\vec{T} \gets \vec{t}, \vec{T}' \gets
\vec{t}', X = x'] \phi$ iff 
$(M,\vec{u}) \sat [\vec{T} \gets \vec{t}, \vec{T}' \gets \vec{t}', X =
x'] \phi$.
\item[SH4.] In context $\vec{u}$, $\Suff - \{X\gets x\}$ depends 
only on $X \gets x$ in $\vec{u}$; that is, for all $\vec{T}'$ disjoint
from $\Suff$ and
$\vec{t}'$, we have $(M,\vec{u}) \sat [\vec{X} \gets x, \vec{T}' \gets
\vec{t}']\Suff$. 
\end{description}
Then 
$X=x$ is a cause of $\phi$ in $(M,\vec{u})$ according to the HP definition.
\ethm

Getting conditions sufficient for causality according to the HP definition
to imply causality according to the NESS test is not so easy.   
The problem is the requirement in the NESS
definition that there be a witness $\Suff$ such that $(M,\vec{u}') \sat
[\Suff] \phi$ in \emph{all} contexts $\vec{u}'$ is very strong, indeed,
arguably too strong.
For example, consider a vote that might be called
off if the weather is bad, where the weather is part of the context.
Thus, in a context where the weather is bad, there is no winner, even if
some votes have been cast.  In the actual
context, the weather is fine and A votes for Mr.~B, who wins the
election.  A's vote is a cause of Mr.~B's victory in this context, according
to the HP definition, but not according to the NESS test, since there is
no set $\Suff$ that includes A sufficient to make Mr.~B win in all
contexts; indeed, there is no cause for Mr.~B's victory according to the
NESS test (which arguably indicates a problem with the definition).

Since the HP definition just focuses on the actual context, 
%for a cause
%of $\phi$ according to the HP definition to be a cause of $\phi$
%according to the NESS test, 
%the variables in $\vec{X} \union \vec{W}$ have to ``screen off'' the
%effect of the context on $\phi$.
%there are no obvious 
%conditions to ensure that, given the settings of $\vec{W}$, 
there is no 
obvious way to conclude from $X=x$ being a cause of $\phi$ in context
$\vec{u}$ a condition holds in all contexts.
To deal with this, I weaken the NESS test so that it must hold only with
respect to a set $U$ of contexts.  More precisely, say that \emph{$\Suff$ is
sufficient for $\phi$ with respect to $U$} if $(M,u) \sat [\Suff]\phi$
for all $u \in U$.  We can then define what it means for $\Suff$ to be
\emph{strongly sufficient for $\phi$ in $(M,\vec{u})$ with respect to
$U$} and for $\vec{X} = \vec{x}$ to be a \emph{cause of $\phi$ in
$(M,\vec{u})$ with respect to $U$} in the obvious way; in the latter
case, we simply  require take strong sufficiency in NT2 and NT3 to be
with respect to $U$.   It is easy to check that
Theorem~\ref{pro:singlecause} holds (with no change in proof) for
causality with respect to a set $U$ of contexts; that is,
even in this case, a cause must be a single conjunct.

\thm\label{pro:HPsuff} Suppose that $X=x$ is a cause of $\phi$ in
$(M,\vec{u})$ according to the HP definition, with $\vec{W}$, $\vec{w}$,
and $x'$ as witnesses.  Suppose that there exists a subset 
$\vec{W}' \subseteq \vec{W}$ such that
%whose values are not changed in $\vec{W}$ (i.e., 
$(M,\vec{u}') \sat \vec{W}' = \vec{w}$ (that is, the assignment
$\vec{W}' \gets \vec{w}$ does not change the values of the variables in
$\vec{W}'$ in context $(M,\vec{u})$)
and a context $\vec{u}'$ such that the following conditions hold, 
where $\vec{W}'' = \vec{W} - \vec{W}'$:
\begin{description}
\item[SN1.] $(M,\vec{u}') \sat [\vec{W}' \gets \vec{w}](X = x' \land
\vec{W}'' = \vec{w})$. 
\item[SN2.] $\vec{W}''$ is independent of $\vec{Z}$ given $X=x$ and
$\vec{W} = \vec{w}$ in
$\vec{u}'$, so that if $\vec{Z}' \subseteq \vec{Z}$, then for all 
$\vec{z}'$, we have
$(M,\vec{u}') \sat [X \gets x, \vec{W}' \gets \vec{w}, \vec{Z}' \gets
\vec{z}'](\vec{W}'' = \vec{w})$.
\item[SN3.] $\phi$ is independent of $\vec{u}$ and $\vec{u}'$
conditional on $X$ and $\vec{W} = \vec{w}$; 
that is if $\vec{Z}' \subseteq \vec{Z}$, then for all $\vec{z}'$
and $x''$, 
we have $(M,\vec{u}') \sat [X \gets x'', \vec{W}
\gets \vec{w}', \vec{Z} \gets \vec{z}'] \phi$ iff 
$(M,\vec{u}) \sat [X \gets x'', \vec{W}
\gets \vec{w}', \vec{Z} \gets \vec{z}'] \phi$.
\end{description}
Then $X=x$ is a cause of $\phi$ in $(M,\vec{u})$
with respect to $\{\vec{u},\vec{u}'\}$ 
according to the causal NESS test.
\ethm

%\end{fullv}

\section{Discussion}\label{sec:conc}
It has long been recognized that normality is a key component of causal
reasoning.  Here I show how it can be incorporated into the HP framework
in a straightforward way.  The HP approach defines causality relative to
a causal model.  But we may be interested in whether a causal statement
follows from some features of the structural equations and some default
statements, without knowing the whole causal model.  For example, in a
scenario with many variables, it may be infeasible (or there might not
be enough information) to provide all the structural equations and a
complete ranking function.  This suggests it
may be of interest to find an appropriate logic for reasoning about actual
causality.  Axioms for causal reasoning (expressed in the language of
this paper, using formulas of the form $[\vec{X} \gets \vec{x}]\phi$,
have already been given by Halpern \citeyear{Hal20}; 
the KLM axioms \cite{KLM} for reasoning about normality and defaults are
well known.  It would be of interest to put these axioms together, 
perhaps incorporating ideas from the causal NESS test, and adding some
statements about (strong) sufficiency, to see if they lead to
interesting conclusions about actual causality.

\paragraph{Acknowledgments:} I thank Steve Sloman for pointing out 
\cite{KM86}, Denis Hilton and Chris Hitchcock for intersting
discussions on causality, and Judea Pearl and the anonymous KR reviewers
for useful comments. 

\journal{
Perhaps the key point of the HP definition is that causality is relative
to a model.  This allows us to tailor the model appropriately.
Suppose that a drunk 18-year-old gets killed in a single-vehicle road
accident.  Many people may focus on the drunkenness as the cause of the
accident.  We all know that you shouldn't drink and drive.  But a road
engineer may want to focus on the too-sharp curve in the road, a
politician may want to focus on the fact that the law allows
18-year-olds to drink, and a psychologist may want to focus on the
youth's recent breakup with his girlfriend.  These different foci wold
be reflected in what we take to be the endogenous and exogenous
variables in the model.  (Recall that it is only endogenous
variables---the ones that can be manipulated---that can be causes.)
To the extent that causality ascriptions are meant to be guides for
future behavior---we ascribe causes so that we know what to do and not
to do next time around---it is perfectly reasonable for different
communities to focus on different aspects of a situation.  Don't drink
and drive; don't build roads with sharp curves; don't allow 18-year-olds
to drink; and don't drive after you've broken up with your girlfriend
may all be useful lessons for different communities to absorb. 
The HP definition allows us to shift focus, by choosing what we take to
be the exogenous and the endogenous variables in the model.

}

%joe
%\bibliographystyle{aaai}
\bibliographystyle{chicagor}
%\bibliography{z,joe,refs}

\nokr{
\appendix 

\section{Appendix: Proofs}
In this appendix, I prove the results stated in the text.  For the
reader's convenience, I repeat the statement of the results here.

\medskip

\opro{pro:singlecause1} If $\vec{X} = \vec{x}$ is a weak cause of
$\phi$ in $(M,\vec{u})$ with $\vec{W}$, $\vec{w}$, and $\vec{x}'$ as
witnesses, $|\vec{X}| > 1$, and each variable  $X_i$ in $\vec{X}$
is independent of all the variables in $\V - \vec{X}$ in $\vec{u}$
(that is, if $\vec{Y} \subseteq \V - \vec{X}$, then for each setting
$\vec{y}$ of $\vec{Y}$, we have
$(M,\vec{u}) \sat \vec{X} = \vec{x}$ iff $(M,\vec{u}) \sat [\vec{Y} \gets
\vec{y}] (\vec{X} = \vec{x})$)

, 
then $\vec{X} =
\vec{x}$ is not a cause of $\phi$ in $(M,\vec{u})$. 
\eopro

\prf  Suppose that the hypotheses of the proposition hold.
First note that since $\vec{X} = \vec{x}$ is a weak
cause of $\phi$ in $(M,\vec{u})$, by AC1, we must have 
$(M,\vec{u}) \sat \vec{X} = \vec{x}$.  Since each variable in
$\vec{X}$ is independent of all the variables in $\V - \vec{X}$, 
for all $\vec{Y} \subseteq \V - \vec{X}$ and all settings $\vec{y}$ of
the variables in $\vec{Y}$, we must have
$(M,\vec{u}) \sat [\vec{Y} \gets \vec{y}](\vec{X} = \vec{x})$.
It follows that, for all formulas $\psi$, all subsets $\vec{X}'$ of
$\vec{X}$, all subsets $\vec{Y}$ of $\V - \vec{X}$, and all settings
$\vec{y}$ of $\vec{Y}$, we have
\begin{equation}\label{eq1}
(M,\vec{u}) \sat [\vec{Y} \gets \vec{y}]\psi \mbox{ iff }
(M,\vec{u}) \sat [\vec{X}' \gets \vec{x}, \vec{Y} \gets \vec{y}]\psi.
\end{equation}

Next, observe that since the causal model is acyclic, there must be some
variable in $\vec{X}$ that is independent of every other variable in
$\vec{X}$.  Without loss of generality, suppose that it is $X_1$.  
Thus, $X_1$ is independent of every variable in $\V - \{X_1\}$. 
Let $\vec{X}^- = \<X_2, \ldots, X_k\>$.
I show that either $X_1 = x_1$ or $\vec{X}^- = \vec{x}$
is a weak cause of $\phi$, showing that $\vec{X} =
\vec{x}$ is not a cause of $\phi$, since it does not satisfy AC3.  

First suppose that $x_1 = x_1'$.  I show that then
$\vec{X}^- = \vec{x}$ is a weak cause of $\phi$, with $\vec{W}$,
$\vec{w}$, and $\vec{x}'$ as witnesses.  To see this, note that since
$\vec{X} =
\vec{x}$ is a weak cause of $\phi$, with $\vec{W}$, $\vec{w}$, and
$\vec{x}'$ as witnesses, by AC2(a), we have that
$(M,\vec{u}) \sat [\vec{X} \gets \vec{x}', \vec{W}
\gets \vec{w}]\neg \phi$. 
By the same arguments as used to derive (\ref{eq1}), we have 
that $(M,\vec{u}) \sat [\vec{X}^- \gets \vec{x}, \vec{W} \gets \vec{w}]
\neg \phi$.  Thus, AC2(a) holds for $\vec{X}^- = \vec{x}$.  By AC2(b), 
$(M,\vec{u}) \sat [\vec{X} \gets \vec{x}, \vec{W}' \gets \vec{w},
\vec{Z}' \gets \vec{z}^*]\phi$ for 
all subsets $\vec{W}'$ of $\vec{W}$ and all subsets $\vec{Z}'$ of
$\vec{Z}$.  By (\ref{eq1}), we have that 
$(M,\vec{u}) \sat [\vec{X}^- \gets \vec{x}, \vec{W}' \gets \vec{w},
\vec{Z}' \gets \vec{z}^*]\phi$.  Thus, AC2(b) holds for  $\vec{X}^- =
\vec{x}$, and $\vec{X}^- = \vec{x}$ is indeed a weak cause of $\phi$.

Now suppose that $x_1 \ne x_1'$.   If $\vec{X}^- = \vec{x}$ is a weak
cause of $\phi$ with witnesses $\vec{W} \union \{X_1\}$, $\vec{w}\cdot
\<x_1'\>$, and $\vec{x}$, then 
we are done.  So suppose that $\vec{X}^- = \vec{x}$ is not a weak
cause of $\phi$ with witnesses $\vec{W} \union \{X_1\}$, $\vec{w}\cdot
\<x_1'\>$, and $\vec{x}$.  It is immediate that AC1 holds for $\vec{X}^- =
\vec{x}$, and that AC2(a) hold with these witnesses.  Thus, AC2(b)
must fail.  It follows that
there must exist some subset $\vec{W}'$ of
$\vec{W}$ and subset $\vec{Z}'$ of $\vec{Z}$ such that either 
(a) $(M,\vec{u}) \sat [\vec{X}^- \gets
\vec{x}, X_1 \gets x_1', \vec{W}' \gets \vec{w},  \vec{Z}' \gets
\vec{z}^*]\neg \phi$ or 
(b) $(M,\vec{u}) \sat [\vec{X}^- \gets
\vec{x}, \vec{W}' \gets \vec{w},  \vec{Z}' \gets
\vec{z}^*]\neg \phi$.  Option (b) cannot hold, because, by 
(\ref{eq1}), it holds iff $(M,\vec{u}) \sat [\vec{X} \gets
\vec{x}, \vec{W}' \gets \vec{w},  \vec{Z}' \gets
\vec{z}^*]\neg \phi$, which contradicts the assumption that $\vec{X}
\gets \vec{x}$ is a weak cause of $\phi$ with $\vec{W}$, $\vec{w}$, and
$\vec{x}'$ as witnesses.  Thus, (a) must hold.
But now it follows that $X_1 = x_1$ is a cause of $\phi$, 
with $\vec{W} \union \vec{X}^-$, $\vec{w} \cdot \vec{x}$, and $x_1'$
as witnesses: AC1 and AC3 are immediate, AC2(a) follows from the
assumption that (a) holds, and AC2(b) follows from the fact
that $\vec{X} = \vec{x}$ is a weak cause with $\vec{W}$, $\vec{w}$, and
$\vec{x}'$ as witnesses.
\eprf

\othm{pro:singlecause} If $\{X_1 = x_1, \ldots, X_k = x_k\}$ is a
cause of $\phi$ in $M$ 
according to the causal NESS test, then $k=1$.
\eothm

\prf Suppose that $\Suff$ is a witness of $\{X_1 = x_1, \ldots, X_k =
x_k\}$ being a cause of 
$\phi$ in $(M,\vec{u})$ according to the causal NESS test and, by way of
contradiction, 
that $k > 1$.  
$\Suff$ is not a witness for $\{X_1 = 1, \ldots, X_{k-1} =
x_{k-1}\}$ being a cause of $\phi$ (otherwise NT4 would be violated).
Thus, it must be the case that $\Suff' = \Suff - \{X_1 = 1, \ldots, X_{k-1} =
x_{k-1}\}$ 
is strongly sufficient for $\phi$ in $(M,\vec{u})$.  But then it follows
that that $X_k = 
x_k$ is a cause of $\phi$ in $(M,\vec{u})$ with $\Suff'$ as a witness.
To see this, note that clearly $\Suff'$ 
satisfies NT1, 
since $\Suff$ does. By assumption, $\Suff'$ is strongly sufficient for
$\phi$ in $(M,\vec{u})$, so NT2 holds.  And, also by
assumption, $\Suff' - \{X_k = x_k\} = \Suff - \{X_1 = x_1, \ldots, X_k =
x_k\}$ is not a strongly sufficient cause of $\phi$, so NT3 holds.  NT4
trivially holds.  This
shows that $\vec{X} = \vec{x}$ is not a cause of $\phi$ according to the
causal NESS test, since it does not satisfy NT4. \eprf

\medskip

\othm{pro:NESSsuff} Suppose that $X=x$ is a cause of $\phi$ in
$(M,\vec{u})$ according 
to the causal NESS test with witness $\Suff$, and there exists a
(possible empty) set $T$ of variables not mentioned in $\phi$ or $\Suff$
and a context $\vec{u}'$ such that the following properties hold:
\begin{description}
\item[SH1.] $\Suff - \{X=x\}$ is not a sufficient condition for $\phi$
in $(M,\vec{u}')$; that is, $(M,\vec{u}') \sat [\Suff - \{X=x\}] \neg \phi$.
\item[SH2.] The variables in $\vec{T}$ depend only on the context
in $\vec{u}$ and $\vec{u}'$; that is, for all $\vec{t}$, $\vec{T}'$
disjoint from $\vec{T}$, and $\vec{t}'$,
we have $(M,\vec{u}) \sat \vec{T} = \vec{t}$ iff $(M,\vec{u}) \sat
[\vec{T}' \gets \vec{t}'] (\vec{T} = \vec{t})$, and similarly for
context $\vec{u}'$.
\item[SH3.] $\phi$ is determined by $\vec{T}$ and $X$ in contexts
$\vec{u}$ and $\vec{u}'$; that is, for all $\vec{t}$, $\vec{T}'$
disjoint from $\vec{T}$ and $X$, $x'$, and $\vec{t}'$,
we have  $(M,\vec{u}') \sat [\vec{T} \gets \vec{t}, \vec{T}' \gets
\vec{t}', X = x'] \phi$ iff 
$(M,\vec{u}) \sat [\vec{T} \gets \vec{t}, \vec{T}' \gets \vec{t}', X =
x'] \phi$.
\item[SH4.] In context $\vec{u}$, $\Suff - \{X\gets x\}$ depends 
only on $X \gets x$ in $\vec{u}$; that is, for all $\vec{T}'$ disjoint
from $\Suff$ and
$\vec{t}'$, we have $(M,\vec{u}) \sat [\vec{X} \gets x; \vec{T}' \gets
\vec{t}']\Suff$. 
\end{description}
Then 
$X=x$ is a cause of $\phi$ in $(M,\vec{u})$ according to the HP definition.
\eothm

\prf Suppose that the hypothesis of the proposition holds.
By SH1, $(M,\vec{u}') \sat [\Suff - \{X=x\}] \neg \phi$.
Choose $x'$ such that
$(M,\vec{u}') \sat [\Suff - \{X=x\}](X=x')$.  I claim that we must have
$x\ne x'$.  For if $(M,\vec{u}') \sat [\Suff - \{X=x\}](X=x)$, then
$(M,\vec{u}') \sat [\Suff]\neg \phi$, contradicting the
assumption that $\Suff$ is strongly sufficient for $\phi$.  Let
$\vec{W}$ consist of all the variables in $\Suff$ other
than $X$, together with the set $\vec{T}$ that satisfies SH2 and SH3;
let $\vec{Z}$ consist of all the remaining endogenous variables.
%other than $X$ whose value is determined by $\vec{u}'$.  
%variables that depend only on $X$ (recursively), and the
%variables in $\phi$.
%variables. 
Let $\vec{w}$ be such that 
$(M,\vec{u}') \sat [\Suff - \{X=x\}](\vec{W} = \vec{w})$. 
Note that $\vec{W} = \vec{w}$ subsumes (i.e., includes all the
assignments in) $\Suff - \{X=x\}$.
It follows that $(M,\vec{u}') \sat [X \gets x', \vec{W}
\gets \vec{w}]\neg \phi$.  
By SH3, we must have  $(M,\vec{u}) \sat [X \gets x', \vec{W} \gets
\vec{w}]\neg \phi$.  Thus, AC2(a) holds.  For AC2(b), let $\vec{W}'$ be
an arbitrary subset of $\vec{W}$ and let $\vec{Z}'$ be an arbitrary
subset of $\vec{Z}$.  As in the statement of AC2(b), suppose that
$(M, \vec{u}) \sat \vec{Z} = \vec{z}^*$.  We want to show that
$(M, \vec{u}) \sat [X \gets x, \vec{W}' \gets \vec{w}, \vec{Z}' \gets
\vec{z}^*]\phi$.  Let $\vec{T}^* = \vec{T} - \vec{W}'$.  Suppose that
$(M,\vec{u}) \sat \vec{T}^* = \vec{t}^*$.  
First note that
since $\Suff$  is strongly sufficient for $\phi$ in $(M,\vec{u})$, we
must have $(M,\vec{u}') \sat [\Suff; \vec{Z}' \gets
\vec{z}^*, \vec{T}^* \gets \vec{t}^*]\phi$.
Let $\vec{W}'' = \vec{W}' \inter \vec{T}$.  Since $\vec{W}'' \subseteq
\vec{T}$ and $(M,\vec{u}') \sat [\Suff - \{X=x\}](\vec{W}'' =
\vec{w})$, by SH2 we must have $(M,\vec{u}') \sat \vec{W}'' = \vec{w}$ and 
$(M,\vec{u}') \sat [\Suff, \vec{Z}' \gets
\vec{z}^*, \vec{T}^* \gets \vec{t}^*]( \vec{W}'' \gets \vec{w})$.
Thus, $(M,\vec{u}') \sat [\Suff, \vec{W}'' \gets \vec{w}, \vec{Z}' \gets
\vec{z}^*, \vec{T}^* \gets \vec{t}^* ]\phi$.
Note that all the variables in $\vec{W}' - \vec{W}''$ are in $\Suff -
\{X\gets x\}$, and they are assigned the same values in $\vec{W}' =
\vec{w}$ as in $\Suff$.  Thus, it follows that 
$(M,\vec{u}') \sat [\Suff, \vec{W}' \gets \vec{w}, \vec{Z}' \gets
\vec{z}^*, \vec{T}^* \gets \vec{t}^*]\phi$.
By SH3, 
$(M,\vec{u}) \sat [\Suff, \vec{W}' \gets \vec{w}, \vec{Z}' \gets
\vec{z}^*, \vec{T}^* \gets \vec{t}^*]\phi$.
By SH2, it follows that $(M,\vec{u}) \sat [\Suff, \vec{W}' \gets
\vec{w}, \vec{Z}' \gets \vec{z}^*]\phi$. Finally, by SH4, 
it follows that $(M,\vec{u}) \sat [X \gets x, \vec{Z}' \gets
\vec{z}^*, \vec{W} \gets \vec{w}]\phi$, as desired.
\eprf

\othm{pro:HPsuff} 
Suppose that $X=x$ is a cause of $\phi$ in
$(M,\vec{u})$ according to the HP definition, with $\vec{W}$, $\vec{w}$,
and $x'$ as witnesses.  Suppose that there exists a subset 
$\vec{W}' \subseteq \vec{W}$ such that
%whose values are not changed in $\vec{W}$ (i.e., 
$(M,\vec{u}') \sat \vec{W}' = \vec{w}$ (that is, the assignment
$\vec{W}' \gets \vec{w}$ does not change the values of the variables in
$\vec{W}'$ in context $(M,\vec{u})$)
and a context $\vec{u}'$ such that the following conditions hold, 
where $\vec{W}'' = \vec{W} - \vec{W}'$:
\begin{description}
\item[SN1.] $(M,\vec{u}') \sat [\vec{W}' \gets \vec{w}](X = x' \land
\vec{W}'' = \vec{w})$. 
\item[SN2.] $\vec{W}''$ is independent of $\vec{Z}$ given $X=x$ and
$\vec{W} = \vec{w}$ in
$\vec{u}'$, so that if $\vec{Z}' \subseteq \vec{Z}$, then for all 
$\vec{z}'$, we have
$(M,\vec{u}') \sat [X \gets x, \vec{W}' \gets \vec{w}, \vec{Z}' \gets
\vec{z}'](\vec{W}'' = \vec{w})$.
\item[SN3.] $\phi$ is independent of $\vec{u}$ and $\vec{u}'$
conditional on $X$ and $\vec{W} = \vec{w}$; 
that is if $\vec{Z}' \subseteq \vec{Z}$, then for all $\vec{z}'$
and $x''$, 
we have $(M,\vec{u}') \sat [X \gets x'', \vec{W}
\gets \vec{w}', \vec{Z} \gets \vec{z}'] \phi$ iff 
$(M,\vec{u}) \sat [X \gets x'', \vec{W}
\gets \vec{w}', \vec{Z} \gets \vec{z}'] \phi$.
\end{description}
Then $X=x$ is a cause of $\phi$ in $(M,\vec{u})$
with respect to $\{\vec{u},\vec{u}'\}$ 
according to the causal NESS test.
\eothm

\prf Let $\Suff = \{X =x, \vec{W}' = \vec{w}\}$.  Clearly NT1 holds.
By assumption, $(M,\vec{u}) \sat [X = x', \vec{W} = \vec{w}] \neg
\phi$.  By SN3, $(M,\vec{u}') \sat [X = x', \vec{W} = \vec{w}] \neg
\phi$.  By SN1, it follows that $(M,\vec{u}') \sat [\vec{W}' =
\vec{w}]\neg \phi$, so NT3 holds.  For NT2, we must show that for all
$\vec{Z}' \subseteq \vec{Z} \union \vec{W}''$, 
%clearly
%$(M,\vec{u}) \sat [X \gets x,\vec{W}' \gets \vec{w}, \vec{Z}' \gets
%\vec{z}^*] \phi$, since $(M,\vec{u}) \sat X = x \land \vec{W}' =
%\vec{w} \land \vec{Z}' = \vec{z}^* \land \phi$.  Moreover, 
%by AC2(b), 
$(M,\vec{u}) \sat [X=x,\vec{W}' = \vec{w}, \vec{Z}' \gets \vec{z}^*]
\phi$, and similarly for $\vec{u}'$.  For $\vec{u}$, this is immediate
from AC2(b).  To see that it also holds for $\vec{u}'$, 
first note that by AC2(b), we also have 
$(M,\vec{u}) \sat [X=x,\vec{W} = \vec{w}, \vec{Z}'' \gets \vec{z}^*]
\sat \phi$, where $\vec{Z}'' = \vec{Z}' \inter \vec{Z}$.
By SN3,
$(M,\vec{u}') \sat [X \gets x,\vec{W} \gets \vec{w}, 
\vec{Z}'' \gets \vec{z}^*] \phi$.  By SN2, it follows that 
$(M,\vec{u}') \sat [X \gets x,\vec{W}' \gets \vec{w}, \vec{Z}' \gets
\vec{z}^*] \phi$.  Thus, NT2 holds with respect to
$\{\vec{u},\vec{u}'\}$.  Clearly NT4 holds, so $X=x$ is a cause of
$\phi$ in $(M,\vec{u})$ with respect to $\{\vec{u},\vec{u}'\}$
according to the causal NESS test.
\eprf
}

%\end{journal}

\journal{\bibliographystyle{chicago}}
\end{document}